\documentclass[useAMS,usenatbib]{coin}
\def\bSig\mathbf{\Sigma}

\providecommand{\tabularnewline}{\\}

\newcommand{\TT}[1]{\mbox{$\tt{#1}$}}
\newcommand{\ORGN}{\mbox{$\tt{ORGN}$}}
\newcommand{\TRUEPDB}{\mbox{$\tt{TRUE}$}}
\newcommand{\PURE}{\mbox{$\tt{PURE}$}}
\newcommand{\EE}{\mbox{$\tt{MTX}$}}

\usepackage[figuresright]{rotating}
\usepackage{color, colortbl}
\usepackage{amsfonts}
\usepackage{longtable}

\usepackage[T1]{fontenc}
\usepackage[latin9]{inputenc}
\usepackage{array}
\usepackage{float}
\usepackage{graphicx}
\usepackage[user,titleref]{zref}
\usepackage{amsmath}
\usepackage{textcomp}
\usepackage{float}
\usepackage{color, colortbl}
\usepackage{multirow}

\usepackage{times} 
\usepackage{helvet}
\usepackage{courier}
\usepackage{verbatim}

\definecolor{LightCyan}{rgb}{0.88,1,1}
\definecolor{LawnGreen}{rgb}{0.48,0.98,0}
\definecolor{LimeGreen}{rgb}{0.19,0.80,0.19}
\definecolor{Red}{rgb}{1,0,0}
\definecolor{LightGrey}{rgb}{0.83,0.83,0.83}
\definecolor{DarkGrey}{rgb}{0.63,0.63,0.63}

\title[A Study of the Effects of Spurious Transitions on Abstraction-based Heuristics]{An Empirical Study of the Effects of Spurious Transitions on Abstraction-based Heuristics}

\author{{\sc Mehdi Sadeqi}\\
        {\it Department of Computer Science, University of Regina}\\
        {\it Regina, Saskatchewan, Canada S4S 0A2}\\
		\and
        {\sc Robert C.\ Holte}\\
        {\it Department of Computing Science, University of Alberta}\\
        {\it Edmonton, Alberta, Canada T6G 2E8}\\
        \and
        {\sc Sandra Zilles}\\
        {\it Department of Computer Science, University of Regina}\\
        {\it Regina, Saskatchewan, Canada S4S 0A2}\\
}

\begin{document}

\setlength\extrarowheight{2pt}
\label{firstpage}
\begin{abstract}
The efficient solution of state space search problems is often attempted by guiding
search algorithms with heuristics (estimates of the distance from any state to the goal). A popular way for creating heuristic functions is by using an abstract version of the state space. However, the quality of abstraction-based heuristic functions, and thus the speed of search, can suffer from spurious transitions, i.e., state transitions in the abstract state space for which no corresponding transitions in the reachable component of the original state space exist. Our first contribution is a quantitative study demonstrating that the harmful effects of spurious transitions on heuristic functions can be substantial, in terms of both the increase in the number of abstract states and the decrease in the heuristic values, which may slow down search. Our second contribution is an empirical study on the benefits of removing a certain kind of spurious transition, namely those that involve states with a pair of mutually exclusive (mutex) variable-value assignments. In the context of state space planning, a mutex pair is a pair of variable-value assignments that does not occur in any reachable state. Detecting mutex pairs is a problem that has been addressed frequently in the planning literature. Our study shows that there are cases in which mutex detection helps to eliminate harmful spurious transitions to a large extent and thus to speed up search substantially.
\end{abstract}


\maketitle
\vspace*{-12pt}

\section{Introduction}
\label{Section-Introduction}

Abstraction is widely used for speeding up search. In particular, it can be beneficial in the creation of heuristic functions. For any state $s$ in the original state space, a heuristic function estimates the distance of $s$ to the goal. Given an abstraction, the actual distances in the abstract space can be used to define a heuristic function that never overestimates the true distances. Such heuristic functions are called admissible; when used to guide search they guarantee that many heuristic search algorithms, like A{*} or IDA{*}, will always find optimal solutions. The key to the efficiency of these algorithms, however, is the quality of the heuristic values. The closer the heuristic values are to the true distances, the more effective they will be in speeding up search.

One problem with abstraction is that it may create ``spurious transitions'' in the abstract space, i.e., abstract transitions between two abstract states with no corresponding pre-image transition\footnote{An abstraction can be given by a mapping from (partial) states to (partial) abstract states, inducing a mapping from operators to ``abstract operators.'' Whenever we refer to the pre-image of an abstract operator or state, we mean the pre-image under that mapping.} in the part of the original state space that is reachable from the start state (original states that are reachable from the start state will also be called genuine original states\footnote{Later in this paper, we sometimes define genuine states with respect to the goal state. In all those cases, all the operators are invertible and therefore the set of states reachable from the goal state equals the set of states reachable from the start state.} in this paper). Similarly, a ``spurious state'' is an abstract state with no corresponding pre-image state in the reachable component of the state space.

Spurious states can be harmful in two ways:\footnote{For a detailed discussion of different types of spurious transitions and how they can be harmful on abstraction-based heuristics, see Section \ref{Section-SpuriousTransitionsAffect}.} (a) they can increase the size of the pattern database (PDB) \citep{DBLP:journals/ci/CulbersonS98}, and (b) by creating shortcuts in the abstract state space, they can produce an optimistic estimate of distances and therefore low quality heuristic values. Consequently, detecting spurious states might result in memory savings and speed up search. While the related literature touches the general issue of spurious transitions and how to avoid at least some of them (see \citep{HelmertAIJ09} and \citep{haslumBG05} for spurious transitions in general and \citep{hernadvolgyiH04}, \citep{haslumBHBK07}, and \citep{zillesH10} for spurious states in particular), there is no work studying the harm these transitions can cause quantitavely. This paper targets this issue. In the meantime, although there is no straightforward approach for detecting all spurious transitions, transitions involving spurious states can be detected by finding the spurious states they connect to. To date, there is no feasible method known for detecting all spurious states. A popular approach to detect some spurious states is to identify those that contain a pair of variable-value assignments that are mutually exclusive (mutex). In the context of state space planning, a mutex pair is a pair of variable-value assignments that does not co-occur in any reachable state. Detecting mutex pairs is a problem that has been addressed frequently in the planning literature, cf.\ Section~\ref{Section-RelatedWork}.

An example of a mutex pair in the standard representation of the Sliding-Tile Puzzle (see Appendix) would be \TT{(TopLeft=} blank and \TT{BottomRight=} blank). Since there is only one blank in this puzzle, two variables cannot simultaneously have the value blank. As another example, consider the Blocks World (see Appendix) problem domain represented using only binary-valued variables of the form ``Block X is on Block Y'',  ``Block X is on the table'', and ``Block X is clear''. The pair of variable-value assignments that say (i) Block 1 is on top of Block 2 and (ii) Block 2 is on top of Block 1 is a mutex pair. 

The following definitions summarize the concepts introduced here. Suppose in what follows that original states are vectors of $n$ variable-value assignments of the form $(x=a)$, where $x$ is a variable and $a$ is one of its possible values; similarly, abstract states are vectors of $m$ variable-value assignments, for some $m\le n$. A partial state is simply the projection of a state (vector) onto some subset of its underlying variables. Further, each abstract state is the output of an abstraction mapping $\Psi$ applied to an original state. The pre-image of any vector $q=((x_1=b_1), \ldots, (x_k=b_k))$ of variable-value assignments is the set of all vectors $p=((x_1=a_1), \ldots, (x_k=a_k))$ of partial original states for which $\Psi(p)=q$. Throughout this document, for simplicity, we assume that every search domain has a single unique goal state $g$.\footnote{Most of our study also applies to the case of multiple goals, as resulting from the use of a goal predicate.} 

\begin{definition}\label{def:spurioustransition} A transition from an abstract state $t$ to an abstract state $t'$ is spurious if there is no pair $(s,s')$ of genuine original states such that (i) $\Psi(s)=t$ and $\Psi(s')=t'$, and (ii) there is a transition from $s$ to $s'$ in the original state space. An abstract state $t$ is spurious if there is no genuine original state $s$ such that $\Psi(s)=t$.
\end{definition}

An abstract state that contains a mutex pair is not necessarily spurious; the abstraction should be taken into consideration before marking an abstract state as spurious. For instance, in the above Blocks World example, consider any projection abstraction that ignores some  variables, while leaving the values of the non-ignored variables unchanged. Then every abstract state containing the mutex pair stating (i) Block 1 is on top of Block 2 and (ii) Block 2 is on top of Block 1, is spurious. By contrast, if the domain is represented with multi-valued variables, some of which represent the names of blocks, and the abstraction makes Block 1 and Block 3 indistinguishable, then this pair should not be considered mutex any longer, as it is the ``abstract image'' of a perfectly valid pair stating that (i) Block 1 is on top of Block 2 and (ii) Block 2 is on top of Block 3. (Another example is given in Subsection \ref{subsec:Type2-Spurious}.) This motivates the following definition.

\begin{definition}\label{def:mutexbased_spuriousstate-0} 
A pair $q$ of variable-value assignments in an abstract state is an \textbf{\textit{abstraction-based mutex pair}} if each pair in the pre-image of $q$ is mutex (in other words, if no pair in the pre-image of $q$ is reachable). A \textbf{\textit{mutex-based spurious state}} is a spurious abstract state that contains an abstraction-based mutex pair.
\end{definition}

To determine whether an abstract state $t$ is a mutex-based spurious state, one applies the abstraction to each pair considered reachable by a mutex detection method. If $t$ contains a pair that is not contained in the resulting list of pairs, then $t$ will be considered a mutex-based spurious state. Note that this method might miss mutex-based spurious states, if the underlying method for finding mutex pairs flags actual mutex pairs as reachable; it might consider non-spurious states spurious, if the underlying method for finding mutex pairs flags reachable pairs as mutex.

Though mutex detection can be an effective tool for removing spurious states, one should be aware of the fact that not every spurious state contains an abstraction-based mutex pair. A problem domain involving constraints on a set of more than two variables may result in ``higher order mutex sets'', i.e., sets of three or more variable-value assignments\footnote{By analogy with the definition of abstraction-based mutex pairs, a tuple $q$ of variable-value assignments in an abstract state is an \emph{abstraction-based mutex set} if all the tuples in the pre-image of $q$ are a ``mutex set'', i.e., not reachable.}. For example, the set of variable-value assignments that say (i) Block 1 is on top of Block 2, (ii) Block 2 is on top of Block 3, and (iii) Block 3 is on top of Block 1, is a mutex set of order 3 in the Blocks World, and thus every abstract state containing this abstraction-based mutex set is spurious. However, this set of assignments does not contain a mutex pair, since every two of the assignments together are valid when ignoring the third one. In fact, every spurious state is itself a higher order mutex, and thus one approach for removing spurious states could be to identify and remove states that contain abstraction-based mutexes of some order. Unfortunately, even for detecting mutex pairs there are no known efficient perfect methods, so that, not surprisingly, there is no known approach for detecting all higher order mutexes. 

Since there are no known efficient methods for detecting \emph{all\/} mutex pairs, existing algorithms usually make a compromise in the number of detected mutex pairs for the computational complexity of the algorithm. Various methods differ in the number and type of mutexes they detect. The first contribution of this work is our systematic empirical study on the effect of spurious transitions on the quality of heuristics. This study is separated by a study on the harm of two types of spurious transitions introduced later. Our second contribution is an empirical study on the benefits of removing those spurious transitions that involve mutex-based spurious states. In this evaluation, we show that there are cases in which mutex detection helps to eliminate harmful spurious transitions to a large extent and thus to speed up search substantially. 

\section[]{Related Work}
\label{Section-RelatedWork}

Spurious states were discussed by Hern{\'a}dv{\"o}lgyi and Holte~(\citeyear{hernadvolgyiH04}) and by Haslum \emph{et al.}\/~(\citeyear{haslumBHBK07}). A systematic theoretical study by Zilles and Holte~(\citeyear{zillesH10}) analyzes the computational complexity of finding abstractions that do not contain any spurious states at all. Although the literature suggests that spurious transitions can have a damaging effect on the size of an abstraction and its corresponding heuristic quality, there is no empirical study demonstrating the intensity of these effects.

Mutex detection was originally suggested in the planning community as the process of finding pairwise conflicts between actions and facts~\citep{Blum95fastplanning}. In the Graphplan~\citep{Blum95fastplanning} planner, two actions or facts are considered mutex if there is no valid plan that could possibly contain the two actions or can possibly make the two facts true in the same time step. An inherent part of almost all planning systems, mutex detection has long been in use for search space pruning and to improve planner performance~\citep{Kautz96pushingthe, Gerevini:2003:PTS:1622452.1622462, DBLP:conf/aaai/PenberthyW94, Vidal:2006:BPO:1141151.1644641, EdelkampH01, Helmert06:0, GereviniS98, Scholz00extractingstate, Rintanen00, BonetG99, FoxL98, haslumBG05, DBLP:journals/ai/ChenHXZ09}. During the search, mutex pairs are enforced for pruning the search space. Mutex detection methods usually achieve efficiency at the cost of missing some mutexes. For example, since Bonet and Geffner's (\citeyear{BonetG99}) algorithm works on grounded representations, in order to make it practical, the search for mutex pairs is systematically limited to a restricted class. The algorithm by Gerevini and Schubert (\citeyear{GereviniS98}) generates many classes of invariants in addition to mutexes, but at the cost of decreasing the performance. An invariant is a property that holds true in all reachable states of a state space. Used in different contexts and various ways, invariants have proven to be useful for finding more mutexes compared to the standard ones extracted from planning graphs.

In the context of abstraction, state-of-the-art techniques for detecting mutexes in the description of an abstract state in binary domain representation, e.g., constrained abstraction \citep{haslumBG05}, usually use particular types of invariants to remove some of the spurious transitions from an abstraction. Occurring quite frequently in well-known benchmark planning domains, mutex pairs are also discussed by \cite{haslumBG05} as a special case of ``at-most-one'' invariants consisting of only two atoms. Haslum~(\citeyear{haslum2006admissible}) also introduces the $h^2$ heuristic, which is a state-of-the-art method for finding mutex pairs. The $h^2$ algorithm for mutex pair detection will be discussed in more detail in Section~\ref{$h^2$-Mutex-Detection}.

\citeauthor{DBLP:conf/ijcai/AlcazarBFF13} (\citeyear{DBLP:conf/ijcai/AlcazarBFF13}) discussed regression planning, which deals with state-sets instead of complete states. They studied reachability-based heuristics and the relationship between mutexes and $e$-deletion (they define spurious states as those states that are generated in a backward search while not being reachable in a forward search). For regression planning, \citeauthor{DBLP:conf/aips/AlcazarT15} (\citeyear{DBLP:conf/aips/AlcazarT15}) showed the impact of invariants and how to compute them. In addition, they suggested using invariants for simplifying the planning task in a pre-processing step 
and observed a noticeable improvement in different planners with no or little drawbacks. In the context of symbolic search, \citeauthor{DBLP:conf/socs/TorralbaA13} (\citeyear{DBLP:conf/socs/TorralbaA13}) showed how to exploit mutexes and invariant groups in conjunction with BDDs. They investigated the pruning power of mutexes found by the $h^2$ algorithm in symbolic search and showed that backward and forward symbolic searches perform similarly well when $h^2$ mutexes are used.

\section{How Spurious Transitions Affect Heuristics} 
\label{Section-SpuriousTransitionsAffect} 

In this section we discuss spurious transitions in detail and quantitatively study their harmful effect on abstraction-based heuristics. The notion of spurious transition is best illustrated with the help of some examples. In these examples, as well as in all our experiments, domains are represented using \emph{production system vector notation} (PSVN)~\citep{hernadvolgyiH99}. The experimented problem domains along with their different representations are described in Appendix. Spurious transitions and spurious states, however, are not a result of using PSVN and can happen in all typical problem definition languages. 

A state in PSVN~\citep{hernadvolgyiH99} is represented by a fixed length vector of characters from the finite alphabet $\Sigma$. Each operator in the operator set has a left and a right hand side, both of them are represented by a fixed length vector equal to the state length. Besides characters from the alphabet, these vectors might also contain some variables and underscore symbols. The left hand side acts as a precondition for an operator. An operator is applicable to a state if the left hand side of an operator can be matched to a state. The variables of the left hand side will be unified to constant characters of the matching state which in turn is used for finding the successor state using the right hand side of the applying operator. Unlike ordinary variables, an underscore in the left hand side means ignoring the character at that position in the state representation. A variable on the right hand side, on the other hand, should be replaced by the unified value of the variable on the left hand side. An underscore on the right hand side means that the character constant of the matched state at that position should be left untouched. The following example illustrates this process. 

Assume the following operator definition with $\Sigma=\{1,2,3,4,5\}$: 

\[\langle A,\_,B,2,\_,B,3 \rangle\rightarrow\langle 3,\_,A,\_,\_,A,B \rangle\]

This operator is only applicable to the states that have a 2 and 3 at the $4^{th}$ and $7^{th}$ positions respectively and also the $3^{rd}$ and $6^{th}$ elements should be the same. For example, $\langle 4,5,1,2,5,1,3 \rangle$ matches the left hand side of this operator which results in binding the variables $A$ and $B$ to 4 and 1 respectively. Considering the definition of the operator and the value of the bound variables, the right hand side will be $\langle3,5,4,2,5,4,1\rangle$. 

Recall that spurious transitions are abstract transitions between two abstract states with no corresponding pre-image transition in the part of the original state space that is reachable from the start state. We distinguish two types of spurious transition:

\begin{description}
\item[Type 1.] Spurious transitions between two non-spurious abstract states.
\item[Type 2.] All other spurious transitions---either connecting a non-spurious abstract state to a spurious abstract state or connecting a spurious abstract state to another spurious abstract state. Note that we are only interested in those spurious states that are reachable from the abstract start state since these are the ones that can increase PDB size and decrease heuristic values. 
\end{description}

\subsection{Example 1: Type 1 Spurious Transitions}

To explain this type of spurious transition, we use an abstraction in the stack representation of Towers of Hanoi (see Appendix). Consider the Towers of Hanoi with 5 disks and 3 pegs in this representation, together with an abstraction $\Psi$ that identifies disks 1, 3, and 5 with each other and disks 2 and 4 with each other. We further assume that the abstract state space is generated by applying this abstraction mapping to the goal state and to all operators, then expanding the abstract goal state using the abstract versions of the inverse operators and iterating this expansion procedure until no new abstract states are produced from the existing ones any more. While this way of generating an abstract state space is common practice, we will show now that it may introduce spurious transitions of type 1.

\begin{figure}[ht]
\noindent \begin{centering}
\includegraphics[scale=0.35]{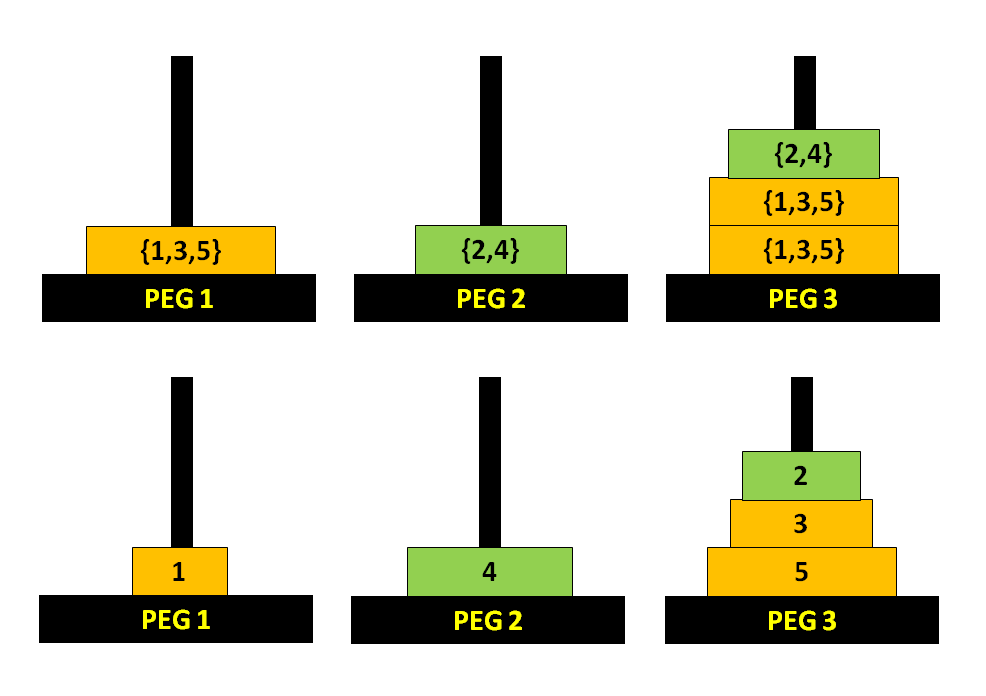} 
\par\end{centering}

\caption{Top: The abstract state $t=\langle 1,1,0,0,0,0;\ 1,2,0,0,0,0;\ 3,1,1,2,0,0 \rangle$ in the stack representation of Towers of Hanoi. The abstraction identifies disks 1, 3, and 5 with each other and disks 2 and 4 with each other. Bottom: The only corresponding reachable original state $s=\langle 1,1,0,0,0,0;\ 1,4,0,0,0,0;\ 3,5,3,2,0,0 \rangle$.}

\label{fig:SpuriousTrans1}
\end{figure}

In the given abstraction, the abstract state \[t=\langle 1,1,0,0,0,0;\ 1,2,0,0,0,0;\ 3,1,1,2,0,0 \rangle\] corresponds to only one reachable original state\footnote{Since all the operators are invertible, the set of states reachable from the start state is equal to the set of states reachable from the goal state.}, namely \[s=\langle 1,1,0,0,0,0;\ 1,4,0,0,0,0;\ 3,5,3,2,0,0 \rangle\,,\] in which disk 1 is on peg 1, disk 4 is on peg 2 and disks 5, 3, and 2 are on peg 3. In other words, $\Psi(s)=t$ and there is no reachable state $s'\ne s$ with $\Psi(s')=t$. $t$ and $s$ are shown in the upper and lower part of Figure \ref{fig:SpuriousTrans1} respectively. 

\begin{figure}[ht]
\noindent \begin{centering}
\includegraphics[scale=0.35]{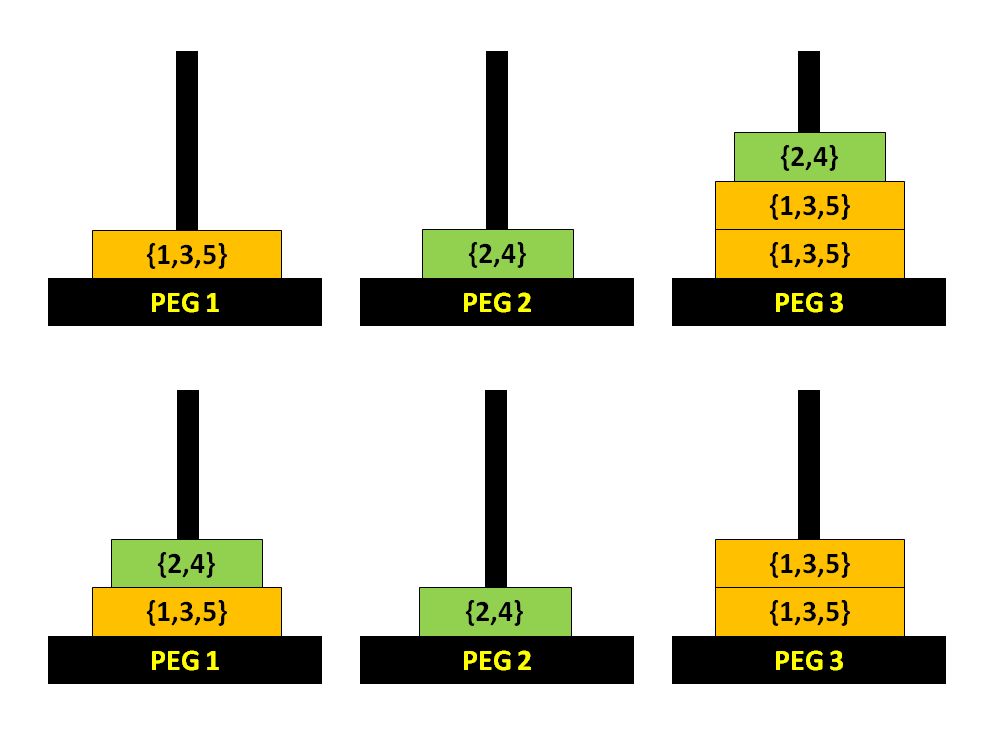} 
\par\end{centering}

\caption{Top: The abstract state $t=\langle 1,1,0,0,0,0;\ 1,2,0,0,0,0;\ 3,1,1,2,0,0 \rangle$ in the stack representation of Towers of Hanoi. Bottom: The abstract state $\Psi(o)(t)$ generated by applying the abstract image of the operator $o$ to the abstract state $t$.}

\label{fig:SpuriousTrans2}
\end{figure}

Consider the operator $o$ given by
\begin{eqnarray*}
&\langle 1,5,0,-,-,-;\ -,-,-,-,-,-;\ 3,-,3,2,-,-\rangle&\\ &\longrightarrow&\\ &\langle 2,5,2,-,-,-;\ -,-,-,-,-,-;\ 2,-,3,0,-,- \rangle&
\end{eqnarray*}
in the definition of the original state space, as well as its abstract image $\Psi(o)$, given by
\begin{eqnarray*}
&\langle 1,1,0,-,-,-;\ -,-,-,-,-,-;\ 3,-,1,2,-,-\rangle&\\ &\longrightarrow&\\ &\langle 2,1,2,-,-,-;\ -,-,-,-,-,-;\ 2,-,1,0,-,- \rangle\,.&
\end{eqnarray*}
Although the original operator $o$ is not applicable to the original state $s$, it turns out that its abstract image $\Psi(o)$ is applicable to the abstract state $\Psi(s)=t$ producing the abstract state
\[\Psi(o)(t)=\langle 2,1,2,0,0,0;\ 1,2,0,0,0,0;\ 3,1,1,0,0,0 \rangle\,.\] 
This is illustrated in Figure \ref{fig:SpuriousTrans2}.

\begin{figure}[ht]
\noindent \begin{centering}
\includegraphics[scale=0.35]{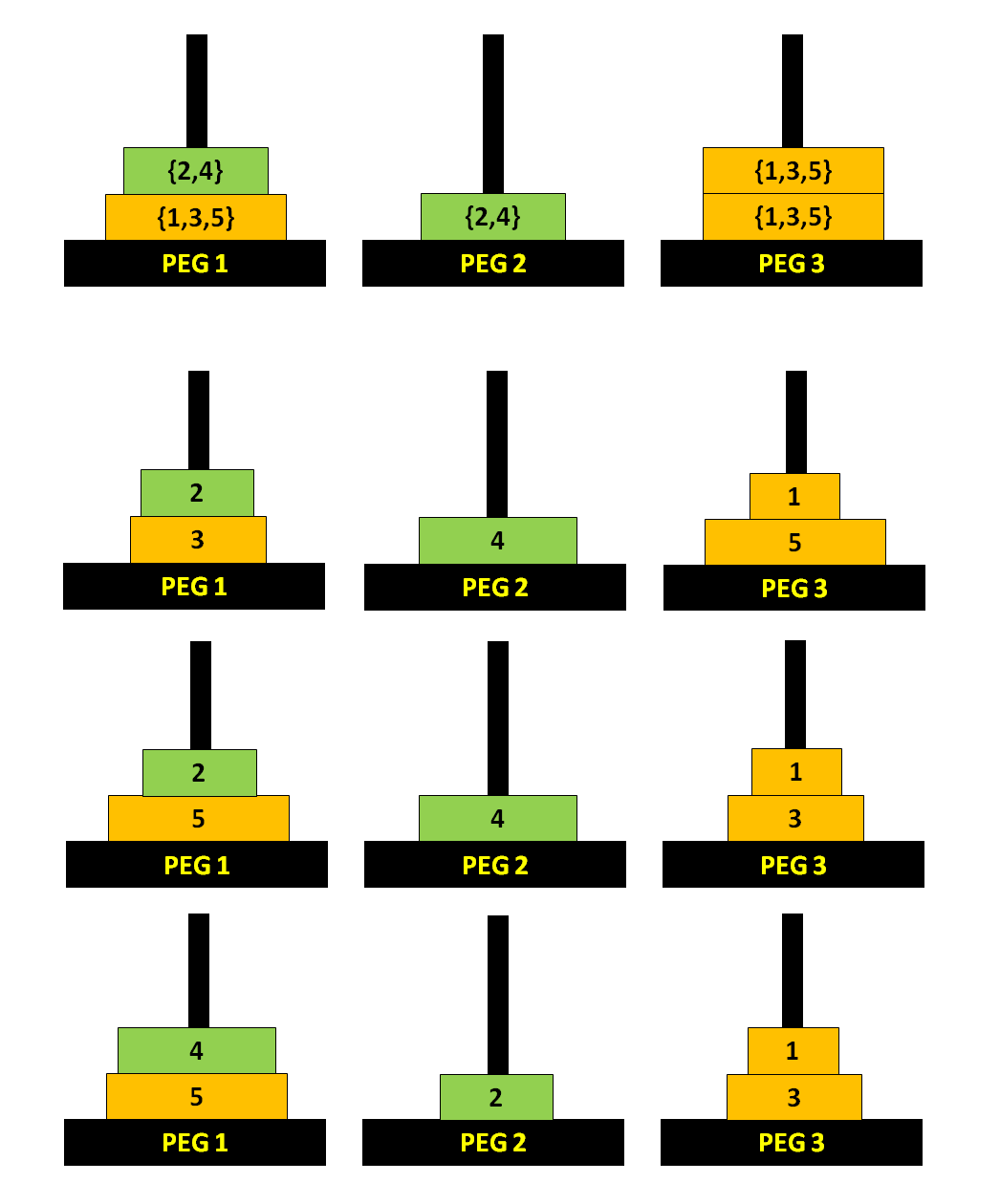} 
\par\end{centering}

\caption{The abstract state $\Psi(o)(t)$, followed by the three reachable states that map to it.}

\label{fig:SpuriousTrans3}
\end{figure}

To show that the transition from $t$ to $\Psi(o)(t)$ in the abstract state space is spurious, we need to show that (i) the pre-image of $t$ contains a genuine state and (ii) there is no operator that maps any genuine state in the pre-image of $t$ to any genuine state in the pre-image of $\Psi(o)(t)$. Statement (i) has already been shown above---the genuine state $s$ is in the pre-image of $t$. To see that statement (ii) is true as well, note first that $s$ is the only genuine state in the pre-image of $t$, so we only have to show that no operator can map $s$ to any state in the pre-image of $\Psi(o)(t)$. There are exactly three genuine states in the pre-image of $\Psi(o)(t)$, as depicted in Figure~\ref{fig:SpuriousTrans3}. For an operator to map $s$ to any of these three states, it would have to move more than one disk at a time (disk 1 moves from peg 1 to peg 3, but also two other disks move to peg 1), which is not possible. Hence we have identified a spurious transition.

Finally, to verify that this spurious transition is of type 1, we have to show that the abstract state $\Psi(o)(t)$ is not spurious, i.e., that its pre-image contains at least one reachable state. However, we have shown above already that there are three reachable states in the pre-image of $\Psi(o)(t)$.

It should be pointed out that the operator definition plays an important role in this example. In the left hand side of the operator $o$, there are some implicit preconditions that can be inferred from the explicitly stated preconditions. For example, the only possible bottom disk of the third peg is disk 4, without this being stated explicitly. Likewise, there is only one option for the height and content of the middle peg: its height is 1 and the only disk on it must be disk 1. Adding these facts, the original operator $o$, given by 
\begin{eqnarray*}
&\langle 1,5,0,-,-,-;\ -,-,-,-,-,-;\ 3,-,3,2,-,-\rangle&\\ &\longrightarrow&\\ &\langle 2,5,2,-,-,-;\ -,-,-,-,-,-;\ 2,-,3,0,-,- \rangle&
\end{eqnarray*}
can be rewritten as
\begin{eqnarray*}
&\langle 1,5,0,-,-,-;\ 1,1,-,-,-,-;\ 3,4,3,2,-,-\rangle&\\ &\longrightarrow&\\ &\langle 2,5,2,-,-,-;\ 1,1,-,-,-,-;\ 2,4,3,0,-,- \rangle&
\end{eqnarray*} 
When this new operator is abstracted it no longer applies to the abstract state $t$ and thus the above-mentioned spurious transition will not be generated. 

In this particular example, it is obvious that this operator is only applicable when disk 4 is at the bottom of peg 3 and disk 1 is at the bottom of peg 2 and we can avoid the spurious transition by explicating these implied preconditions in the operator. Though we can enumerate these types of implied preconditions for small size domains, it is not obvious that there exists an efficient method for bigger size domains. Including \emph{all the implied preconditions} for a similar operator of a bigger size domain may require the enumeration of many combinations of disks leading to an exponential number of operators. In an arbitrary problem domain and in the extreme case one might have to explicitly enumerate all the states in operators to prevent type 1 spurious transitions. This means that, in general, if there are an exponential number of disjunctive implied preconditions then we currently have no better option than to leave them implicit, opening ourselves up to type 1 spurious transitions. This also emphasizes the importance of the representation and the crucial fact that even a very subtle change in how an operator is encoded can be the difference between having and not having spurious transitions.

\subsection{Example 2: Type 2 Spurious Transitions Increasing the PDB Size} \label{subsec:Type2-Spurious}

\begin{figure}[ht]
\noindent \begin{centering}
\includegraphics[scale=0.8]{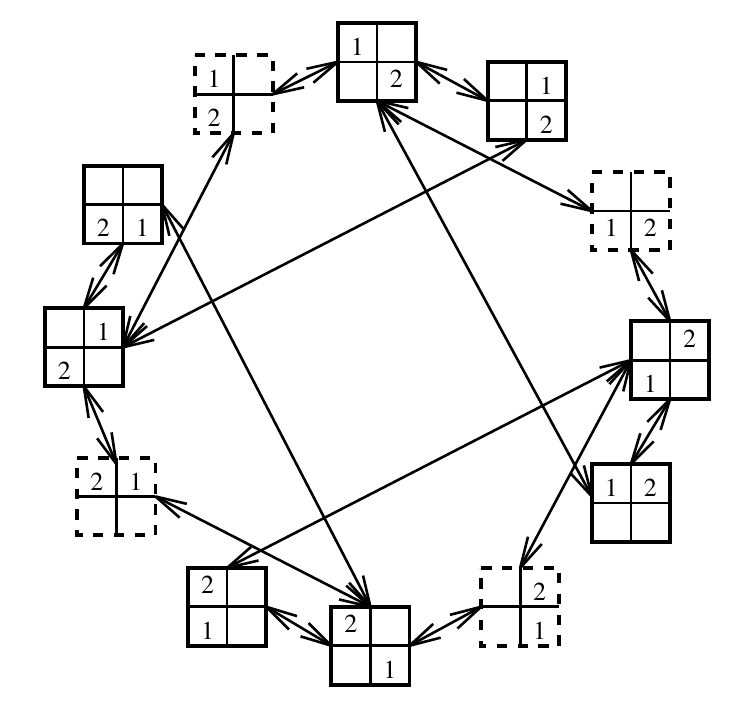}
\par\end{centering}

\caption{Abstraction of the $2\times2$ Sliding-Tile Puzzle in standard representation,
adopted from~\citep{hernadvolgyiH04}. Spurious states represented by dashed boxes increase the PDB size.}

\label{fig:spurious1}
\end{figure}

To see how an abstraction can create spurious transitions connecting a non-spurious state to a spurious one, consider the standard representation of the $2\times2$ Sliding-Tile Puzzle under an abstraction created by mapping every occurrence of tile $3$ to a $B$. Having two blanks allows more moves from any given state. Starting at the abstract state $\langle1,2,B,B\rangle$ (the image of the goal state $\langle1,2,3,B\rangle$), some move sequences reach abstract states that do not correspond to any reachable state in the original space---spurious states. Figure \ref{fig:spurious1} shows the abstract space; every solid box represents an abstract state reachable from the abstract image of the goal state, every dashed box represents a spurious state---in the original state space reachable from the original goal state, there is no state that maps to it. In this figure, every arrow to a spurious state represents a type 2 spurious transition.

In Figure~\ref{fig:spurious1}, a mutex pair would be a pair of variable assignments that says that tile $1$ is in the upper left corner and tile $2$ is in the bottom left corner---there is no reachable original state that, in the particular abstraction displayed there, has this variable assignment in its abstract state. Since the abstract state displayed in the 11 o'clock position contains this abstraction-based mutex pair, it must be a mutex-based spurious state and can be deleted from the PDB.

We already mentioned that an abstract state containing a mutex pair is not necessarily spurious and the abstraction should be taken into consideration before marking an abstract state as spurious. To better understand this, consider the abstraction of Example 2. Since there is only one blank in this puzzle, having two blanks at each two locations will be a mutex pair. If we do not consider the abstraction, each abstract state in Figure \ref{fig:spurious1} should be considered spurious. This, however, is not true. We should notice that none of these two blanks in each of the abstract states is a mapping from two actual blanks in the original space, i.e., one of these blanks in each of these abstract states is always a $3$ in the original space. In other words, for two blanks in the each abstract state, as long a $3$ and a blank appear in the corresponding locations in the original space, this will not be a real mutex pair and does not necessarily make that abstract state spurious.

The example shown in Figure \ref{fig:spurious1} does not feature any type 1 spurious transitions. Furthermore, the present spurious transitions of type 2 (i.e., the spurious states) increase the PDB size, but they do not create shortcuts and thus do not decrease the heuristic values. A decrease in heuristic values occurs in larger versions of the puzzle or in some abstractions in other representations of this puzzle. A different example best illustrates the decrease in the heuristic quality potentially caused by spurious transitions.

\subsection{Example 3: Type 2 Spurious Transitions Increasing the PDB Size and Decreasing the Heuristic Quality}

\begin{figure}[ht]
\noindent \begin{centering}
\includegraphics[scale=0.8]{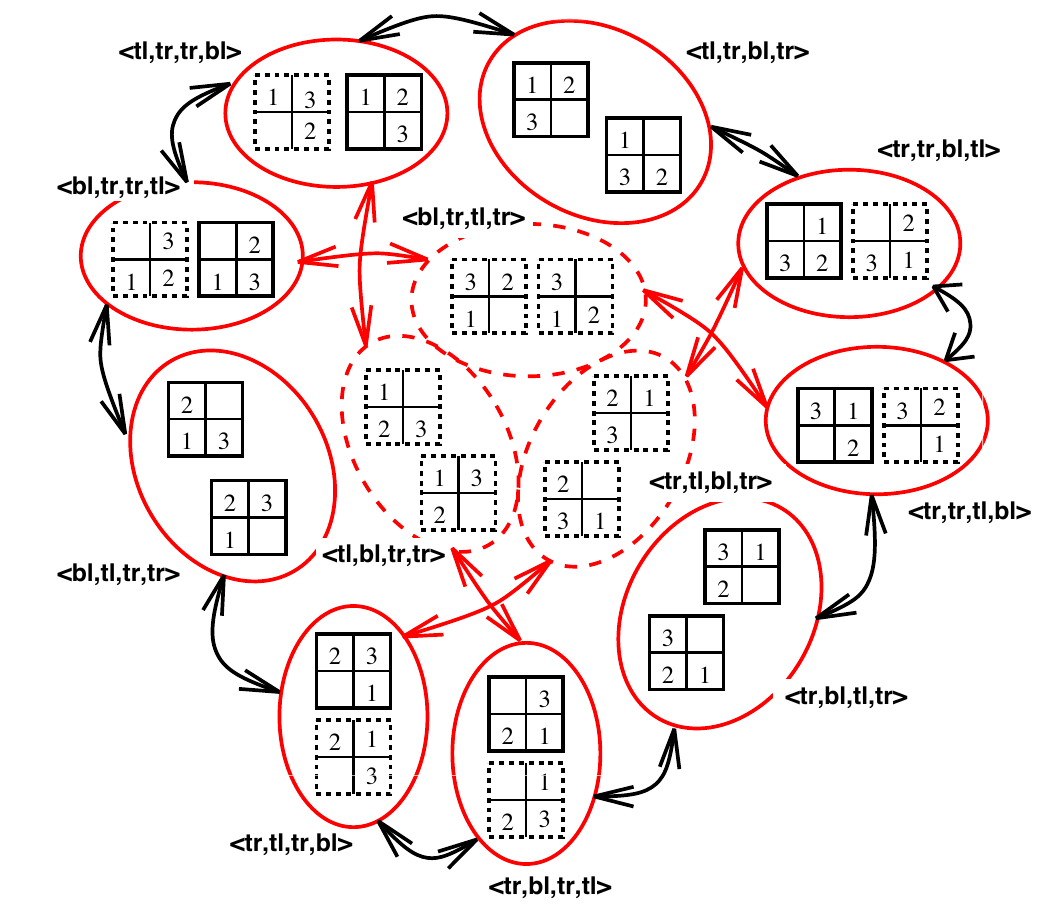}
\par\end{centering}

\caption{Abstraction of the $2\times2$ Sliding-Tile Puzzle in dual representation, adopted from~\citep{hernadvolgyiH04}.
In this scenario, spurious states increase the PDB size and decrease the heuristic quality.}

\label{fig:spurious2}
\end{figure}

For this type of spurious transition, we use an example abstraction in the dual representation of the $2\times2$ Sliding-Tile Puzzle (see Appendix for a description of this representation). In Figure~\ref{fig:spurious2}, the position numbers are replaced by location names: \emph{bl} for \emph{bottom left}, \emph{tl} for \emph{top left}, \emph{tr} for \emph{top right}. The abstraction shown identifies the bottom right location with the top right location. Ellipses show how original states are grouped by the abstraction; dashed ellipses are spurious states since they contain only unreachable original states (those in dashed boxes). The abstract state $\langle bl,tr,tr,tl\rangle$ is not spurious, because the reachable original state $\langle bl,tr,br,tl\rangle$ maps to it (see the solid box inside the ellipse). Similar to the previous example, all arrows pointing to a dashed ellipse are spurious transitions of type 2.

Here one can see that the spurious transitions create shortcuts in the abstract space: the distance between $\langle bl,tr,tr,tl\rangle$ and $\langle tr,tr,tl,bl\rangle$ is 2, but would be 4 if the spurious state $\langle bl,tr,tl,tr\rangle$ was removed.

This example shows that spurious states, and thus spurious transitions of type 2, can decrease heuristic values, but it is not obvious that the effect on the quality of the heuristic can be noticeable in terms of search time (for larger domains), even after mutex-based spurious states have been removed. Our goal is to provide empirical evidence for this statement. For that purpose, we compare abstractions that contain spurious states to their versions after the application of mutex filters and after removing all spurious states, and thus all spurious transitions of type 2. As in the previous example, all spurious transitions occurring in the present example are of type 2. In Subsection \ref{sub-type1-effect}, we will also show in some detail how type 1 spurious transitions can substantially decrease the heuristic quality.

Before going into the details of the experiments, it is required to discuss the abstraction selection procedure. In every case, the abstractions were chosen manually so that the corresponding abstract state spaces have certain interesting properties (e.g., containing spurious states, containing mutex-based spurious states or containing type 1 spurious transitions, etc.) and also small enough to be computable in terms of the required time and memory. 

\section{Mutex Pair Detection for Nuetralizing the Effects of Spurious Transitions} \label{sec:MutexDetection}

In this section we study the effects of mutex pair detection in neutralizing the effects of type 2 spurious transitions. We start by a formal definition of a mutex pair and continue with an experimental study of the effect of removing all type 2 spurious transitions containing a mutex pair in some small state spaces. The method used for detecting all mutexes is not scalable to bigger size domains and therefore we have to use a more efficient mutex detection method for this purpose. A state-of-the-art mutex detection method, $h^2$, is chosen for these situations.

\subsection{Formal Definition of a Mutex Pair} \label{sec:ExhaustiveMutex-0}

Assume that states are represented as variable-value pairs in $n$ variables. We denote the variables with $x_1,\ldots,x_n$, so that the state vector $(a_1,\ldots,a_n)$ corresponds to the assignment vector $(x_1=a_1,\ldots,x_n=a_n)$. Propositional logic variables, such as those commonly used in planning, are treated as variables that can take on one of two values (true and false). With this convention, we formally define reachable state and mutex pair.

\begin{definition}
Suppose $s^*$ is any fixed state. We call a state reachable if it is reachable from $s^*$. For any $i,j$ with $1\le i<j\le m$ and any $a_i,a_j$, the partial original state $(x_i=a_i,x_j=a_j)$ is a reachable pair if there are $a_k$, for $k\in\{1,\ldots,m\}\setminus\{i,j\}$ such that $(a_1,\ldots,a_m)$ is a reachable state; otherwise $(x_i=a_i,x_j=a_j)$ is a mutex pair. (Note that, since mutex pairs are defined with respect to $s^*$, the latter has to be chosen carefully. In particular, $s^*$ should be chosen such that the goal state is reachable from it.) 
\end{definition}

\subsection{Exhaustive Mutex Pair Detection} \label{sec:ExhaustiveMutex}

We will first focus on detecting type 2 spurious transitions, i.e., spurious states. Our first set of experiments is run on state spaces small enough that the states and all mutex pairs can be enumerated exhaustively and the true abstraction---containing no spurious states---can be computed.

For the first set of experiments, we chose the Blocks World with 9 blocks and 3 table positions, with domain abstraction applied to the top representation and projection applied to the height representation;  the $3\times 4$ Sliding-Tile Puzzle with projection applied to both the standard and the dual representation; the Towers of Hanoi with 9 disks and 4 pegs with domain abstraction applied to the stack representation\footnote{In this representation, domain abstractions that map non-consecutive disks will result in abstractions containing spurious states \citep{hernadvolgyiH04}.}; the 6-Belt Scanalyzer with projection, and the Constrained-Movement Sliding-Tile Puzzle in sizes $3\times 4$ and $4\times 5$ with domain abstraction. (See Appendix for a description of the domains and their representations.) For each representation we chose several abstractions, for a total of 66 abstractions. Throughout this document, we chose the abstractions such that the corresponding abstract state spaces have certain properties of interest such as containing spurious states, mutex-based spurious states or type 1 spurious transitions. They also needed to be small enough to be computable with respect to the required time and memory.

For each abstraction we compare three PDBs based on that abstraction, namely (i)~\ORGN, the original PDB containing spurious states\footnote{For these experiments, the spurious states are those generated from the abstract goal state using abstract inverse operators.}, (ii)~\EE, the PDB produced by removing from \ORGN\ all abstract states containing an abstraction-based mutex pair and (iii)~\TRUEPDB, the PDB produced by removing all spurious states from \ORGN.

We then evaluate each abstraction in terms of (i) the sizes of \ORGN\ and \EE\ compared to the size of \TRUEPDB, and (ii) the number of nodes expanded by IDA{*} using either of \ORGN\ or \EE\ compared to the number of nodes expanded using \TRUEPDB.  To compare the number of nodes expanded, say for \ORGN\ to that for \TRUEPDB, we first sample 1,000 start states uniformly at random, compute the ratio of the number of nodes expanded using \ORGN\ over the number of nodes expanded when using \TRUEPDB, and then compute the average over the 1,000 obtained ratios.\footnote{Note that averaging the ratios of two tuples of numbers $A=(a_1,...,a_n)$ and $B=(b_1,...,b_n)$ where $n \in \mathbb{N}$, only makes sense when either (i) $a_i \leq b_i$ for all $ i \in \{1,...,n\}$, or (ii) $a_i \geq b_i$ for all $ i \in \{1,...,n\}$; or informally, one tuple ``dominates'' the other tuple. In our experiments, this has always been the case when we have reported this measure. We have chosen the average of ratios over the ratio of averages for two main reasons: first, we needed to decrease the effect of extreme cases or outliers; and, second, we were interested in measuring the improvement achieved for every problem instance independently. In addition, since the problem instances were chosen independently, the arithmetic average seemed to be a more suitable option than the geometric average.} The same set of randomly sampled states is used for the comparison of \EE\ to \TRUEPDB.

We do not measure the size of the PDB in terms of the memory used but in terms of the number of states in the PDB. Our rationale for doing so is that the data structures used for storing PDBs vary. While the memory used for storing a PDB does not necessarily grow with the number of states in the PDB, for some often used and well scalable implementations, this is the case. The reader is referred to Section~\ref{sec:pdb} for a more detailed discussion on this issue.

\begin{figure}[ht]
\noindent \begin{centering}
\includegraphics[scale=0.63]{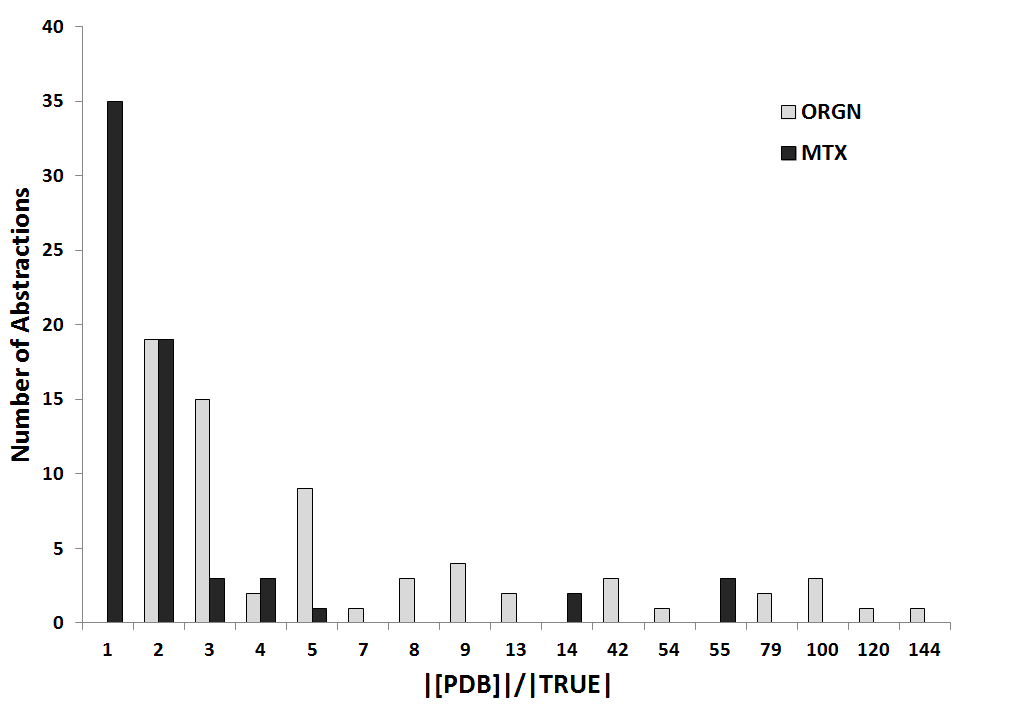}

\par\end{centering}

\caption{$x$-axis: Ratio of the sizes of \ORGN\ and \EE\ to the size of \TRUEPDB. This axis is not linear scale and shows only the numbers that occur in our experiments. $y$-axis: Number of abstractions for the corresponding $x$-value. The light-coloured and dark bars are for \ORGN\ and \EE, respectively.}

\label{fig:PDBRatioEE}
\end{figure}

Figure \ref{fig:PDBRatioEE} shows the histogram of relative sizes of \ORGN\ and \EE\ with respect to the size of \TRUEPDB. The $x$-axis represents the ratio of the sizes of \ORGN\ and \EE\ to the size of \TRUEPDB. They respectively illustrate how many spurious states are created by an abstraction and how many of these spurious states have mutex pairs. Each number $n$ on this axis represents the range $(n-1,n]$ of values. The $y$-axis shows the number of abstractions (out of 66) for the corresponding $x$-value. The light-coloured bars represent this number for \ORGN, while the dark bars are for \EE.

For example, the light-coloured bar at the $x$-value of $100$ has a height of $3$ indicating that $3$ out of the $66$ abstractions we studied have between 98 and 99 times as many spurious states as non-spurious states. For $47$ of our $66$ abstractions, \ORGN\ is twice the size of \TRUEPDB\ or more. Also, as shown by the dark bar at the $x$-value of $1$, in a total of $35$ of the $66$ abstractions we tried, all the spurious states have been eliminated by mutex pair detection, i.e., all spurious states are mutex-based. In the remaining cases, some spurious states did not contain abstraction-based mutex pairs and thus could not be eliminated using mutex pair detection. In all our experiments, there were some spurious states that contained abstraction-based mutex pairs. Overall, in the abstractions we tested, there is a strong tendency of the dark bars towards small $x$-values, suggesting a high  effectiveness of exhaustive mutex pair detection in eliminating spurious states.

\begin{figure}[ht]
\noindent \begin{centering}
\includegraphics[scale=0.36]{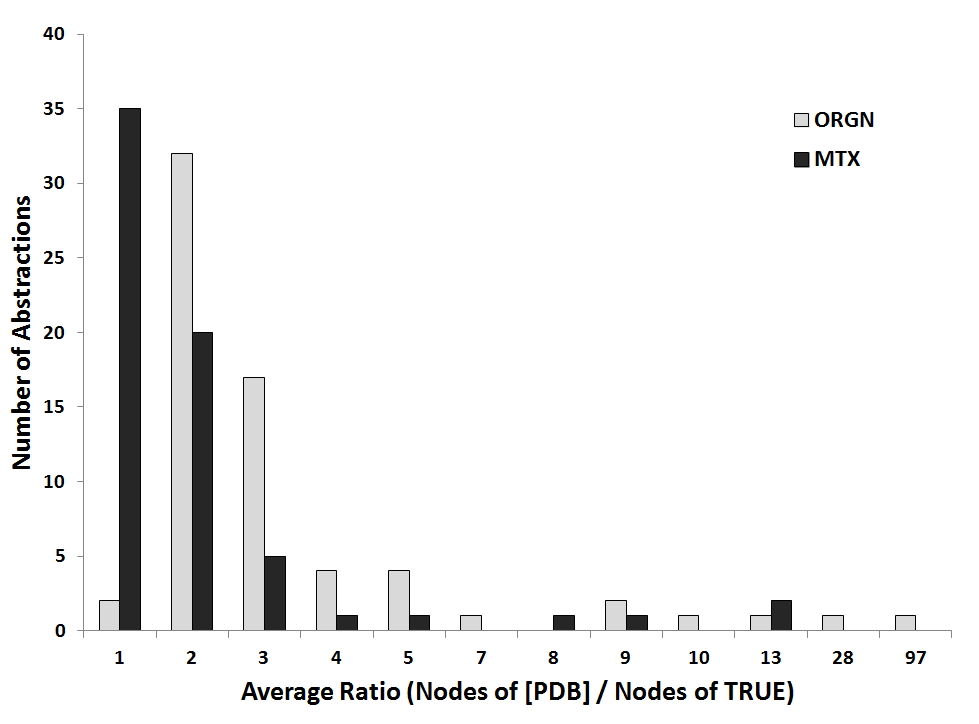}
\par\end{centering}

\caption{$x$-axis: Average ratio of the number of nodes expanded by IDA* of \ORGN\ and \EE\ to the corresponding number of \TRUEPDB. The $x$-axis is not linear scale and only the numbers that have an occurrence in our experiments are shown here. $y$-axis: Number of abstractions for the corresponding $x$-value with the light-coloured bar for \ORGN\ and the dark bar for \EE.}

\label{fig:NodesExpandedRatioEE}
\end{figure}

Figure~\ref{fig:NodesExpandedRatioEE} is similar to Figure~\ref{fig:PDBRatioEE} but is based on the number of nodes expanded by IDA* in solving problems, not on the number of abstract states in the PDB.
The $x$-axis represents the average ratio of the number of nodes expanded by IDA* using \ORGN\ and \EE\ over the number of nodes using \TRUEPDB. Each number $n$ on this axis represents the range $(n-1,n]$ of values. Similar to Figure~\ref{fig:PDBRatioEE}, the $y$-axis shows the number of abstractions for the corresponding $x$-value with the light-coloured bar representing \ORGN\ and the dark bar representing \EE. 

We observe that spurious states often have a strong negative effect on the quality of the heuristic in the abstractions we tested---32 out of our 66 abstractions
have an $x$-value of three or greater (i.e., IDA* takes more than twice as long to solve a problem if the spurious states are present) and 7 of them have an $x$-value greater than $6.58$.
As in Figure~\ref{fig:PDBRatioEE}, the large $y$-value for \EE\ at $x=1$ indicates that, in many cases, mutex pair detection has completely eliminated the quality loss of the heuristic due to spurious states. This does not mean that in all these cases mutex pair detection removes all spurious states---there could be spurious states that do not reduce the heuristic values at all. Interestingly, in 5 cases, mutex pair detection eliminated many spurious states, but the ones it did \emph{not\/} eliminate were those that caused the decrease of the heuristic values. In other words, in some cases only the spurious states that do not contain abstraction-based mutex pairs are harmful in terms of the heuristic quality. As extreme examples of this, in the three tested domain abstractions of the $4\times5$ version of the Constrained-Movement Sliding-Tile Puzzle, we have abstractions having 54,432,000 abstract states, 53,121,600 of which are spurious. While 52,617,600 of these spurious states contain abstraction-based mutex pairs, their removal does not change the resulting average heuristic value and removing the remaining 504,000 spurious states improves the average heuristic value in these cases (see Table \ref{tab:SP-SE-4by5} of Appendix). Another example is one abstraction in the $3\times4$ version of this puzzle having 237,827,520 spurious states of which 236,960,640 contain abstraction-based mutex pairs. Removing these 236,960,640 spurious states does not change heuristic quality but removing the remaining 866,880 spurious states improves the average heuristic value (see the second row of Table \ref{tab:SP-SE-3by4} of Appendix).

A high-level summary of our results is given in Table~\ref{tab:ALLRepResults}. This table divides the abstractions into 4 categories according to two criteria. The first criterion is whether or not the number of spurious states in \ORGN\ is at least as large as the number of non-spurious states (see rows categorized by problem domains and also shown in total). The second criterion is whether or not the number of nodes expanded by IDA* using \ORGN\ was more than twice the number expanded when using \TRUEPDB. If yes, this abstraction falls into the category ``IDA* slow'', otherwise it falls into the category ``IDA* normal''. Each cell in the last two columns of the table shows the total number of abstractions that exhibited the row-column combination of effects for each problem domain and in total.

\begin{table}[h!]
\centering
\begin{tabular}{c|c||c|c|}
\cline{2-4}
\multicolumn{1}{c|}{} & Problem Domain & \textbf{IDA* normal}  & \textbf{IDA* slow} \tabularnewline
\cline{2-4}
\hline
\multicolumn{1}{|c|}{\multirow{5}{*}{\textbf{\# spurious $\ge$ \# non-spurious}}} & BW    & 1  & 7  \tabularnewline
\cline{2-4}
\multicolumn{1}{|c|}{} & STP   & 11 & 11 \tabularnewline
\cline{2-4}
\multicolumn{1}{|c|}{} & ToH   & 4  & 6  \tabularnewline
\cline{2-4}
\multicolumn{1}{|c|}{} & CSTP  & 4  & 2  \tabularnewline
\cline{2-4}
\multicolumn{1}{|c|}{} & SCNZ  & 1  & 0  \tabularnewline
\cline{1-4}
\rowcolor{LightGrey} \multicolumn{1}{|c|}{\textbf{Total}} &  & \textbf{21} & \textbf{26} \tabularnewline
\hline
\hline

\multicolumn{1}{|c|}{\multirow{5}{*}{\textbf{\# spurious $<$ \# non-spurious}}} & BW    & 3  & 5  \tabularnewline
\cline{2-4}
\multicolumn{1}{|c|}{} & STP   & 1 & 0 \tabularnewline
\cline{2-4}
\multicolumn{1}{|c|}{} & ToH   & 1 & 1  \tabularnewline
\cline{2-4}
\multicolumn{1}{|c|}{} & CSTP  & 0  & 0  \tabularnewline
\cline{2-4}
\multicolumn{1}{|c|}{} & SCNZ  & 8  & 0  \tabularnewline
\cline{1-4}
\rowcolor{LightGrey} \multicolumn{1}{|c|}{\textbf{Total}} &  & \textbf{13} & \textbf{6} \tabularnewline
\hline
\end{tabular}

\caption{Abstractions categorized by different types of harm caused by spurious
states (BW: Blocks World, STP: Sliding-Tile Puzzle, ToH: Towers of Hanoi,
CSTP: Constrained-Movement Sliding-Tile Puzzle, SCNZ: Scanalyzer).}

\label{tab:ALLRepResults}
\end{table}

From Table \ref{tab:ALLRepResults}, we observe that spurious states can slow down IDA* substantially (in our experiments, a little less than half the cases as shown in Table \ref{tab:ALLRepResults}) and can also cause PDBs to be much larger than they need to be. In our tested abstractions, mutex pair detection was able to neutralize the negative effects of spurious states to a reasonable extent. This motivates the development of an efficient mutex pair detection method for multi-valued domains. Two such methods are proposed in \citep{SadeqiHZ13-2,DBLP:conf/ai/SadeqiHZ14} and \citep{SadeqiHZ13}.

\subsection{Effect of Scalable Domain-Independent Mutex Pair Detection}

In this section we quantitatively demonstrate the effect of removing mutex pairs on improving the heuristic function quality in bigger size domains. This is done by application of a state-of-the-art mutex detection method, $h^2$, on various domain and projection abstractions of such domains. Before discussing these results we need to know how $h^2$ mutex detection works and how the original PDB is modified by removing mutex pairs detected by $h^2$ in order to create the so-called $h^2$-PDB.

\subsubsection{$h^2$ Mutex Detection} \label{$h^2$-Mutex-Detection}
Most existing mutex detection methods use some form of invariant synthesis in the process of mutex detection. A state-of-the-art mutex detection method, $h^2$, discovers mutex pairs as a special case of ``at-most-one'' invariants consisting of only two atoms.\footnote{For more background on domain analysis and invariants the reader is referred to \citep{haslum2006admissible}. See also Section \ref{Section-RelatedWork}.} The $h^2$ invariant synthesis process can be summarized as follows:

\begin{itemize}
\item The (pairs of) atoms of the initial state (e.g., start state) are reachable.

\item An operator is considered applicable if all single atoms and pairs of atoms in its preconditions are reachable. 

\item An applicable operator makes reachable all of its single add effects and all pairs made in one of the following ways:
\begin{itemize}
		\item from the add effects of the operators,
		\item any add effect combined with any previous
               reachable atom which is not deleted by the operator and is not mutex with one of its preconditions.
\end{itemize}
\end{itemize}

The original PDB is built by moving backwards starting from the abstract goal applying the abstract inverse operators in a breadth-first manner. Creating the $h^2$-modified PDB (called $h^2$-PDB) is quite straightforward. The only difference to the original PDB creation is that while moving backwards from the abstract goal, an abstract state containing an abstraction-based mutex pair is not added to the open-list.

\subsubsection{Effect of $h^2$ Mutex Detection on the Heuristic Function}
The quantitative demonstration of the effect of removing mutex pairs on the heuristic function quality in bigger size domains is done by application of $h^2$ mutex detection on the domain abstractions of the 15-Blocks World with 3 Table Positions in top representation, projection abstractions of the $4\times 5$ Sliding-Tile Puzzle in both the standard and the dual representation and the projection abstractions of the 12-Belt Scanalyzer. Two medium size problem domains from planning, Depot and Storage, are also tested for this demonstration.

For the Blocks World, the Sliding-Tile Puzzle and Scanalyzer, Tables~\ref{tab:BW-TOP-15by3-H2-1} to~\ref{tab:SCNZ-ST-12-Belts-H2-1} show these abstractions, their PDB size and average heuristic value before and after removing mutex pairs using $h^2$, along with the percentage of improvement of heuristic values of the $h^2$-PDB with respect to the original PDB. The rows of these tables are sorted in decreasing order of the size of the original PDB. For the Depot and Storage experiments, Tables~ \ref{tab:DEPOT-FD-10-H2-SelectedAbsts-1} and \ref{tab:STORAGE-FD-13-H2-1} show the experimented abstractions, their PDB size and the average number of nodes expanded by IDA* using the original PDB and $h^2$-PDB. To calculate the average number of nodes expanded, 1,000 uniformly chosen problem instances are solved by IDA* and their number of nodes expanded are averaged. These problem instances were generated by moving forward from the canonical start state in the problem definition file, enumerating all the states of the state space reachable from this start state and uniformly choosing problem instances from the entire set of reachable states. Since the experimented sizes of these two problem domains are small enough to solve all problem instances with the tested abstractions, we have used the more informative measure of the average number of nodes expanded by IDA* instead of the average heuristic value. Again, the rows of these tables are sorted in decreasing order of the size of the original PDB.

The first column illustrates the abstraction rules applied. A rule \[a\gets a_{1},a_{2},\ldots,a_{k}\] means that the symbols $a_{1},a_{2},\ldots,a_{k}$ are no longer distinguishable and are all mapped to the symbol $a$ (domain abstraction). A rule \[\mbox{keep [\emph{facts}]}\] means that the variables encoding the listed facts are kept and the remaining ones are ignored (projection abstraction). For example, in the top representation of Blocks World, ``$2\leftarrow12,13,14$'' describes an abstraction in which blocks $12$, $13$ and $14$ are mapped together. Similarly, in the standard representation of the Sliding-Tile Puzzle, ``keep [locations
$1$, $2$, $5$]'' describes an abstraction in which the variables encoding the puzzle grid locations $1$, $2$ and $5$ are kept and the rest are ignored.

The heuristic value of 1,000 uniformly chosen random problem instances is averaged to obtain the average heuristic value of each PDB, shown in columns 2 and 3. The same 1,000 random problem instances are used for calculation of the percentage of improvement of heuristic values of the $h^2$-PDB over the original PDB, shown in the last column of each table. For every problem instance, the percentage of improvement of the $h^2$-PDB heuristic value with respect to the corresponding number of the original PDB is calculated and the resulting percentages are averaged. In the following subsections, we discuss the results for each domain separately.

\subsubsection{15-Blocks World with 3 Table Positions in Top Representation}
In the tested domain abstractions of the 15-Blocks World with 3 Table Positions in top representation, the heuristic function can be remarkably improved by application of $h^2$ (see Table~\ref{tab:BW-TOP-15by3-H2-1}). In all but one case, the $h^2$ mutex detection improves the heuristic by more than $10\%$. In all abstractions in this table, the heuristic is improved substantially although the number of spurious states removed by $h^2$ is small compared to the number of states in the original PDB. This shows that a small number of spurious states can have a notable effect on the heuristic quality, emphasizing the importance of mutex detection.

\begin{table}[h!]

\begin{footnotesize}

\centerline{\begin{tabular}{|l|l|l|l|}
\hline 
\rowcolor{LightGrey} {Abstraction} & {Original} & {$h^{2}$-PDB} & {\% of improvement}\tabularnewline
\rowcolor{LightGrey} & {Size} & {Size} & \tabularnewline
\rowcolor{LightGrey} & {Avg.\ $h$} & {Avg.\ $h$} & \tabularnewline
\hline
{$0\leftarrow0,1,2,3,4$} & {138,442,746} & {136,809,394} & {1.18} \tabularnewline
{$1\leftarrow5,6,7,8,9,10,11$} & {18.19} & {21.42} & {18.08}\tabularnewline
{$2\leftarrow12,13$} &  &  & \tabularnewline
{$3\leftarrow14,15$} &  &  & \tabularnewline
\hline
{$0\leftarrow0,1,2,3,4$} & {33,006,979} & {32,813,959} & {0.58} \tabularnewline
{$1\leftarrow5,6,7,8,9,10,11,12$} & {17.26} & {18.96} & {9.98}\tabularnewline
{$2\leftarrow13,14$} &  &  & \tabularnewline
\hline
{$0\leftarrow0,1,2,3,4,5,6$} & {20,928,426} & {20,778,306} & {0.72} \tabularnewline
{$1\leftarrow8,9,10,11,12,13$} & {15.27} & {16.81} & {10.20}\tabularnewline
{$2\leftarrow14,15$} &  &  & \tabularnewline
\hline
{$0\leftarrow0,1,2,3,4$} & {4,003,497} & {3,980,619} & {0.57} \tabularnewline
{$1\leftarrow5,6,7,8,9,10,11,12,13$} & {14.13} & {15.94} & {13.07}\tabularnewline
{$2\leftarrow14,15$} &  &  & \tabularnewline
\hline
{$0\leftarrow0,1,2,15$} & {3,179,780} & {3,163,766} & {0.50} \tabularnewline
{$1\leftarrow3,4,5,6,7,8,9,10,11,12$} & {14.40} & {16.30} & {13.45}\tabularnewline
{$2\leftarrow13,14$} &  &  & \tabularnewline
\hline
{$0\leftarrow0,1,2,3,4,5,6,7$} & {2,072,988} & {2,056,974} & {0.77} \tabularnewline
{$1\leftarrow8,9,10,11,12,13$} & {12.56} & {14.01} & {11.65}\tabularnewline
{$2\leftarrow14,15$} &  &  & \tabularnewline
\hline
{$1\leftarrow1,2,3,4,5,6,7,8,9,10,11,12,13$} & {378,547} & {377,429} & {0.29} \tabularnewline
{$2\leftarrow14,15$} & {13.44} & {15.52} & {16.08}\tabularnewline
\hline
\end{tabular}{\footnotesize }}{\footnotesize \par}


\caption{Sample domain abstractions of the top representation of Blocks World with 15 Blocks and 3 Table Positions showing the effect of removing mutex pairs on the heuristic quality and the PDB size using $h^{2}$.}

\label{tab:BW-TOP-15by3-H2-1}

\end{footnotesize}

\end{table}

\subsubsection{$4\times 5$ Sliding-Tile Puzzle in Standard and Dual Representation}

Tables~\ref{tab:SP-ST-4by5-H2-1} and~\ref{tab:SP-DU-4by5-H2-1} show the negative effect of spurious states containing mutexes on some projection abstractions of the $4\times 5$ Sliding-Tile Puzzle in the standard and the dual representation, respectively. In the standard representation, the percentage of improvement is always less than 7\% despite the fact that many spurious states are removed by mutex detection. Compared to the abstractions of 15-Blocks World with 3 Table Positions in top representation, the mutex-based spurious states have a smaller negative effect here. This effect becomes more noticeable in the dual representation. In 4 out of 6 cases, the percentage of improvement is more than 12\% illustrating the importance of removing mutex-based spurious states.


%

\begin{table}[h!]

\begin{footnotesize}

\centerline{\begin{tabular}{|l|l|l|l|}
\hline 
\rowcolor{LightGrey} {Abstraction} & {Original} & {$h^{2}$-PDB} & {\% of improvement}\tabularnewline
\hline 
\rowcolor{LightGrey} & {Size} & {Size} & {Size} \tabularnewline
\rowcolor{LightGrey} & {Avg.\ $h$} & {Avg.\ $h$} & {Avg.\ $h$} \tabularnewline
\hline
{keep [locations 6,10,11,17,18,20]} & {64,000,000} & {27,907,200} & {56.39} \tabularnewline
 & {9.95} & {10.63} & {6.96}\tabularnewline
\hline
{keep [locations 9,13,14,18,19,20]} & {64,000,000} & {27,907,200} & {56.39} \tabularnewline
 & {9.80} & {10.24} & {4.58}\tabularnewline
\hline
{keep [locations 7,8,9,19,20]} & {3,200,000} & {1,860,480} & {41.86} \tabularnewline
 & {8.25} & {8.63} & {4.87}\tabularnewline
\hline
{keep [locations 1,2,12,17,18]} & {3,200,000} & {1,860,480} & {41.86} \tabularnewline
 & {9.07} & {9.17} & {1.08}\tabularnewline
\hline
{keep [locations 9,10,11,17,19]} & {3,200,000} & {1,860,480} & {41.86} \tabularnewline
 & {9.08} & {9.21} & {1.49}\tabularnewline
\hline
{keep [locations 1,2,11,12,20]} & {3,200,000} & {1,860,480} & {41.86} \tabularnewline
 & {8.17} & {8.66} & {6.24}\tabularnewline
\hline
{keep [locations 17,18,19,20]} & {160,000} & {116,280} & {27.32} \tabularnewline
 & {6.27} & {6.46} & {3.24}\tabularnewline
\hline
{keep {[}locations 1,2,19,20]} & {160,000} & {116,280} & {27.32} \tabularnewline
 & {6.34} & {6.55} & {3.48}\tabularnewline
\hline
{keep [locations 1,10,17,18]} & {160,000} & {116,280} & {27.32} \tabularnewline
 & {7.32} & {7.39} & {1.05}\tabularnewline
\hline
\end{tabular}{\footnotesize }}{\footnotesize \par}


\caption{Sample projection abstractions of the standard representation of the $4\times5$-Sliding-Tile Puzzle showing the effect of removing mutex pairs on the heuristic quality and the PDB size using $h^{2}$.}

\label{tab:SP-ST-4by5-H2-1}

\end{footnotesize}

\end{table}


%
\begin{table}[h!]

\begin{footnotesize}

\centerline{\begin{tabular}{|l|l|l|l|}
\hline 
\rowcolor{LightGrey} {Abstraction} & {Original} & {$h^{2}$-PDB} & {\% of improvement}\tabularnewline
\hline 
\rowcolor{LightGrey} & {Size} & {Size} & {Size} \tabularnewline
\rowcolor{LightGrey} & {Avg.\ $h$} & {Avg.\ $h$} & {Avg.\ $h$} \tabularnewline
\hline
{keep [tiles 1,2,3,12,13,14]} & {64,000,000} & {27,907,200} & {56.39} \tabularnewline
 & {17.35} & {17.64} & {1.94}\tabularnewline
\hline
{keep [tiles 1,5,6,13,14,blank]} & {64,000,000} & {27,907,200} & {56.39} \tabularnewline
 & {31.85} & {36.14} & {13.84}\tabularnewline
\hline
{keep [tiles 1,8,18,19,blank]} & {3,200,000} & {1,860,480} & {41.86} \tabularnewline
 & {30.98} & {34.77} & {12.50}\tabularnewline
\hline
{keep [tiles 1,5,6,16,17]} & {3,200,000} & {1,860,480} & {41.86} \tabularnewline
 & {15.01} & {15.26} & {2.03}\tabularnewline
\hline
{keep [tiles 16,17,18,19,blank]} & {3,200,000} & {1,860,480} & {41.86} \tabularnewline
 & {27.97} & {33.99} & {22.27}\tabularnewline
\hline
{keep [tiles 17,18,19,blank]} & {160,000} & {116,280} & {27.32} \tabularnewline
 & {23.60} & {27.65} & {17.95}\tabularnewline
\hline
\end{tabular}{\footnotesize }}{\footnotesize \par}


\caption{Sample projection abstractions of the dual representation of the $4\times5$-Sliding Tile Puzzle showing the effect of removing mutex pairs on the heuristic quality and PDB size using $h^{2}$.}

\label{tab:SP-DU-4by5-H2-1}

\end{footnotesize}

\end{table}

\subsubsection{12-Belt Scanalyzer}
Table~\ref{tab:SCNZ-ST-12-Belts-H2-1} shows some projection abstractions of the 12-Belt Scanalyzer that we tested. In these abstractions, the mutex pairs do not cause much harm and therefore removing them with $h^2$ does not have a remarkable effect on improving the heuristic quality\footnote{Because of the minor difference between the average heuristic value of the original and the $h^2$-PDB, they are shown with 3 decimal places in this table.} (performance improvements of at most 1\%). There is also a case in which the mutex-based spurious states have no effect on the heuristic quality and removing them with $h^2$ does not change the heuristic function (the highlighted row). This illustrates that spurious states do not necessarily have a negative effect on the heuristic quality (similar to the $2\times2$-Sliding-Tile Puzzle example 1 in Section \ref{subsec:Type2-Spurious}).

\begin{table}[h!]

\begin{footnotesize}

\centerline{\begin{tabular}{|l|l|l|l|}
\hline 
\rowcolor{LightGrey} {Abstraction} & {Original} & {$h^{2}$-PDB} & {\% of improvement}\tabularnewline
\hline 
\rowcolor{LightGrey} & {Size} & {Size} & {Size} \tabularnewline
\rowcolor{LightGrey} & {Avg.\ $h$} & {Avg.\ $h$} & {Avg.\ $h$} \tabularnewline
\hline
{keep belts 2,3,5,6,9,11} & {95,551,488} & {21,288,960} & {77.72} \tabularnewline
{keep bln\_analyzed 2,5,8,9,11} & {5.676} & {5.734} & {1.00}\tabularnewline
\hline
{keep belts 5,6,7,8} & {42,467,328} & {24,330,240} & {42.71} \tabularnewline
{keep bln\_analyzed 0,1,2,3,4,5,6,7,8,9,11} & {9.069} & {9.095} & {0.27}\tabularnewline
\hline
{keep belts 5,6,9,11} & {42,467,328} & {24,330,240} & {42.71} \tabularnewline
{keep bln\_analyzed 0,2,3,6,9} & {9.069} & {9.095} & {0.27}\tabularnewline
\hline
{keep belts 0,5,7} & {7,077,888} & {5,406,720} & {23.61} \tabularnewline
{keep bln\_analyzed 0,1,2,3,4,5,6,7,8,9,10,11} & {11.855} & {11.906} & {0.44}\tabularnewline
\hline
{keep belts 2,8,9} & {7,077,888} & {5,406,720} & {23.61} \tabularnewline
{keep bln\_analyzed 0,1,2,3,4,5,6,7,8,9,10,11} & {12.305} & {12.312} & {0.06}\tabularnewline
\hline
{keep belts 0,1,2} & {7,077,888} & {5,406,720} & {23.61} \tabularnewline
{keep bln\_analyzed 0,1,2,3,4,5,6,7,8,9,10,11} & {11.799} & {11.803} & {0.03}\tabularnewline
\hline
{keep belts 6,7,8} & {3,538,944} & {2,703,360} & {23.61} \tabularnewline
{keep bln\_analyzed 0,1,2,3,4,5,6,7,8,9,11} & {8.860} & {8.893} & {0.38}\tabularnewline
\hline
{keep belts 2,5,8,11} & {663,552} & {380,160} & {42.71} \tabularnewline
{keep bln\_analyzed 1,4,5,8,10} & {4.107} & {4.128} & {0.53}\tabularnewline
\hline
{keep belts 3,6,9,10} & {663,552} & {380,160} & {42.71} \tabularnewline
{keep bln\_analyzed 2,5,6,9,10} & {4.656} & {4.661} & {0.12}\tabularnewline
\hline
{keep belts 10,11} & {589,824} & {540,672} & {8.33} \tabularnewline
{keep bln\_analyzed 0,1,2,3,4,5,6,7,8,9,10,11} & {11.361} & {11.362} & {0.02}\tabularnewline
\hline
\rowcolor{DarkGrey} {keep belts 0,1} & {589,824} & {540,672} & {8.33} \tabularnewline
\rowcolor{DarkGrey} {keep bln\_analyzed 0,1,2,3,4,5,6,7,8,9,10,11} & {11.334} & {11.334} & {0.00}
\tabularnewline
\hline
{keep belts 3,9,11} & {13,824} & {10,560} & {23.61} \tabularnewline
{keep bln\_analyzed 2,9,11} & {3.300} & {3.305} & {0.18}\tabularnewline
\hline
\end{tabular}{\footnotesize }}{\footnotesize \par}


\caption{Sample projection abstractions of the standard representation of the 12-Belts Scanalyzer showing the effect of removing mutex pairs on the heuristic quality and PDB size using $h^{2}$.}
\label{tab:SCNZ-ST-12-Belts-H2-1}

\end{footnotesize}

\end{table}

\subsubsection{Depot}
The multi-valued representation derived by Fast Downward's \citep{Helmert06:0} preprocessing algorithm from \texttt{pfile10}---in the \texttt{depot} folder of the benchmark problem domains that come with the Fast Downward planner package\footnote{One can obtain the Fast Downward planner package from \texttt{http://www.fast-downward.org}.}---is used for the experiments in this problem domain. Table \ref{tab:DEPOT-FD-10-H2-SelectedAbsts-1} shows some projection abstractions of this representation of the Depot domain that we tested. The average number of nodes expanded by IDA* using the original PDB and $h^2$-PDB are compared. 1,000 uniformly chosen problem instances are selected for this comparison. These problem instances were generated by moving forward from the canonical start state in the problem definition file. Since we have experimented with a small size of this problem domain, we were able to enumerate all the states of the state space reachable from the start state and uniformly chose problem instances from this entire set. The table also compares the two PDBs in terms of the number of abstract states stored.

In all the experiments here, the projection abstractions contain many mutex-based spurious states. However, the detected mutex pairs in these abstractions do not cause much harm on the heuristic quality and therefore removing them with $h^2$ does not have a remarkable effect on the average number of nodes expanded by IDA* (performance improvements of at most 1.17\%). Similar to the Scanalyzer experiments, we even have a case in which the spurious states detected by $h^2$ have no effect on the heuristic quality (the highlighted row). This is another example illustrating that mutex-based spurious states do not necessarily have a negative effect on the heuristic quality as was illustrated in the $2\times2$-Sliding-Tile Puzzle example 1 in Section \ref{subsec:Type2-Spurious}.

\begin{table}[h!]

\begin{footnotesize}

\centerline{\begin{tabular}{|l|l|l|l|}
\hline 
\rowcolor{LightGrey} {Abstraction} & {Original} & {$h^{2}$-PDB} & {\% of improvement}\tabularnewline
\hline 
\rowcolor{LightGrey} & {Size} & {Size} & {Size} \tabularnewline
\rowcolor{LightGrey} & {Avg. Nodes} & {Avg. Nodes} & {Avg. Nodes} \tabularnewline
\hline
{keep {[}2,4,6,7,12,14,15,16,21,25,26,30,31{]}} & {87,945,984} & {31,491,456} & {64.19} \tabularnewline
 & {143,172,898} & {142,801,915} & {0.26}\tabularnewline
\hline
{keep {[}1,25,26,27,29,30,31,32{]}} & {64,117,932} & {26,079,613} & {59.32} \tabularnewline
 & {23,446,300} & {23,179,219} & {1.15}\tabularnewline
\hline
{keep {[}1,26,27,29,30,31,32{]}} & {32,058,966} & {14,580,083} & {54.52} \tabularnewline
 & {23,446,300} & {23,179,219} & {1.15}\tabularnewline
\hline
{keep {[}1,24,27,29,30,31,32{]}} & {32,058,966} & {14,580,083} & {54.52} \tabularnewline
 & {23,580,620} & {23,312,461} & {1.15}\tabularnewline
\hline
{keep {[}2,4,8,9,14,15,20,21,25,26,30,31{]}} & {12,563,712} & {4,873,536} & {61.21} \tabularnewline
 & {125,419,502} & {125,118,004} & {0.24}\tabularnewline
\hline
\rowcolor{DarkGrey} {keep {[}1,2,20,27,28,30,31{]}} & {11,811,198} & {3,948,529} & {66.57} \tabularnewline
\rowcolor{DarkGrey}  & {120,106,291} & {120,106,291} & {0.00}\tabularnewline
\hline
{keep {[}4,9,24,29,30,31,32{]}} & {3,374,628} & {1,780,374} & {47.24} \tabularnewline
 & {42,929,498} & {42,446,576} & {1.14}\tabularnewline
\hline
{keep {[}1,13,21,25,27,30,31,32{]}} & {6,749,256} & {3,186,782} & {52.78} \tabularnewline
 & {105,841,133} & {104,618,281} & {1.17}\tabularnewline
\hline
\end{tabular}{\footnotesize }}{\footnotesize \par}

\caption{Sample projection abstractions of Fast Downward's representation of Depot \texttt{pfile10} showing the effect of removing mutex pairs on the heuristic quality and PDB size using $h^{2}$.}

\label{tab:DEPOT-FD-10-H2-SelectedAbsts-1}

\end{footnotesize}

\end{table}

\subsubsection{Storage}
The multi-valued representation derived by Fast Downward's preprocessing algorithm from \texttt{p13.pddl}---in the \texttt{storage} folder of the benchmark problem domains that come with the Fast Downward planner package---is used for the experiments in this problem domain. Table~\ref{tab:STORAGE-FD-13-H2-1} shows our tested projection abstractions of this representation of the Storage domain, comparing the average number of nodes expanded by IDA* using the original PDB and the $h^2$-PDB. 1,000 uniformly chosen problem instances are selected for this purpose. These problem instances were generated by moving forward from the canonical start state in the problem definition file. Since we have experimented with a small size of this problem domain, we were able to enumerate all the states of the state space reachable from the start state and uniformly chose problem instances from the entire set. The table also compares the two PDBs in terms of the number of abstract states stored.

Similar to the examples from the Depot domain, in all the experiments here, the projection abstractions contain many mutex-based spurious states. The deleted mutex pairs in these abstractions do not cause much harm on the heuristic quality and therefore removing them with $h^2$ does not have a substantial effect on the average number of nodes expanded by IDA* (except for two cases, performance improvements of less than 7\%).

\begin{table}[h!]

\begin{footnotesize}

\centerline{\begin{tabular}{|l|l|l|l|}
\hline 
\rowcolor{LightGrey} {Abstraction} & {Original} & {$h^{2}$-PDB} & {\% of improvement}\tabularnewline
\hline 
\rowcolor{LightGrey} & {Size} & {Size} & {Size} \tabularnewline
\rowcolor{LightGrey} & {Avg. Nodes} & {Avg. Nodes} & {Avg. Nodes} \tabularnewline
\hline
{keep {[}1,3,7,12,18,19,20,24,32,38,40,41,42{]}} & {268,435,456} & {71,368,834} & {73.41} \tabularnewline
 & {2,388,434} & {2,081,803} & {14.73}\tabularnewline
\hline
{keep {[}1,3,5,6,12,24,25,26,38,40,41,42{]}} & {134,217,728} & {75,514,048} & {43.74} \tabularnewline
 & {5,183,681} & {5,106,698} & {1.51}\tabularnewline
\hline
{keep {[}2,11,14,24,25,26,27,30,31,32,34,35,37,38,41,42{]}} & {33,554,432} & {10,018,880} & {70.14} \tabularnewline
 & {4,467,062} & {4,353,486} & {2.61}\tabularnewline
\hline
{keep {[}1,2,3,7,12,18,19,20,24,32,38,40,41{]}} & {33,554,432} & {9,091,012} & {72.91} \tabularnewline
 & {2,243,215} & {2,050,529} & {9.40}\tabularnewline
\hline
{keep {[}1,2,3,5,6,12,24,26,39,40,42{]}} & {8,388,608} & {4,676,844} & {44.25} \tabularnewline
 & {3,099,433} & {3,006,516} & {3.09}\tabularnewline
\hline
{keep {[}1,2,3,4,7,12,18,19,20,24,32,38,41{]}} & {4,194,304} & {1,159,532} & {72.35} \tabularnewline
 & {2,243,215} & {2,098,119} & {6.91}\tabularnewline
\hline
{keep {[}1,5,6,12,24,25,26,38,41,42{]}} & {4,194,304} & {3,035,600} & {27.62} \tabularnewline
 & {5,183,681} & {5,137,187} & {0.90}\tabularnewline
\hline
{keep {[}1,3,4,6,8,9,18,19,24,32,38,42{]}} & {2,097,152} & {1,177,184} & {43.87} \tabularnewline
 & {5,392,829} & {5,377,139} & {0.29}\tabularnewline
\hline
{keep {[}1,30,31,32,33,34,35,36,37,38,42{]}} & {1,048,576} & {277,394} & {73.54} \tabularnewline
 & {3,048,086} & {2,938,303} & {3.74}\tabularnewline
\hline
{keep {[}1,3,7,12,18,19,20,24,32,38,40{]}} & {1,048,576} & {329,836} & {68.54} \tabularnewline
 & {2,444,543} & {2,287,527} & {6.86}\tabularnewline
\hline
{keep {[}30,31,32,33,34,35,36,37,38,42{]}} & {65,536} & {20,089} & {69.35} \tabularnewline
 & {4,594,489} & {4,533,861} & {1.34}\tabularnewline
\hline
\end{tabular}{\footnotesize }}{\footnotesize \par}

\caption{Sample projection abstractions of Fast Downward's Representation of Storage \texttt{p13.pddl} showing the effect of removing mutex pairs on the heuristic quality and PDB size using $h^{2}$.}

\label{tab:STORAGE-FD-13-H2-1}

\end{footnotesize}

\end{table}

\subsection{Type 1 Spurious Transition Detection} \label{sub-type1-effect}

Up to this point, we have been only discussing the effects of type 2 spurious transitions and trying to remove them by detecting spurious states. However, it could be the case that an abstraction also contains type 1 spurious transitions.

To detect type 1 spurious transitions, we have to find the set of all authentic edges and remove any edges that do not belong to this set. When the size of the problem domain is small this can be easily done by traversing the entire reachable component of the original state space and collecting all the edges in this component. For every abstraction we will apply the abstraction on this reachable edge set to obtain the set of reachable abstract edges in the abstract space. While computing the PDB, we use this set of reachable abstract edges to remove the type 1 spurious transitions. 

\begin{figure}[ht]
\noindent \begin{centering}
\includegraphics[scale=0.65]{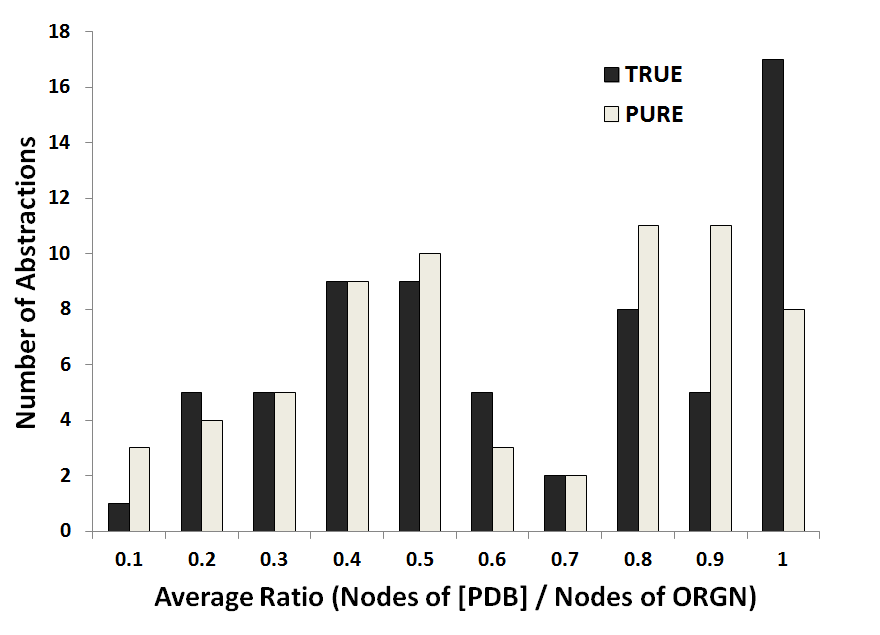}
\par\end{centering}

\caption{$x$-axis: Ratio of the number of nodes expanded by IDA* of \TRUEPDB\ and \PURE\ to the corresponding number of \ORGN. $y$-axis: Number of abstractions for the corresponding $x$-value.}

\label{fig:pureTransitionsGraph}
\end{figure}

Figure \ref{fig:pureTransitionsGraph} shows the histogram of relative numbers of nodes expanded by IDA* after removing all spurious transitions (both type 1 and type 2) compared to the corresponding number after removing all spurious states. The $x$-axis is the ratio of the number of nodes expanded by IDA* of \TRUEPDB\ and \PURE\ to the corresponding number of \ORGN. \PURE\ is the PDB produced by removing both type 1 and type 2 spurious transitions from \ORGN. Every number $n$ on this axis represents a range of values of $(n-0.1,n]$. Similar to previous histograms in the current section, the $y$-axis shows the number of abstractions for the corresponding $x$-value. The light-coloured bars are for the ratio of the number of nodes expanded by IDA* for the \PURE\ PDBs to the number of nodes expanded using the original PDBs containing all spurious transitions. The dark bars represent the corresponding numbers for the PDBs after removing only spurious states. As can be seen in this histogram, the light-coloured bars are closer to 0 illustrating the improvement after removing type 1 spurious transitions.


In our experiments, type 1 spurious transitions did not happen for domain abstractions of the top representation of the Blocks World but they are sometimes present in projection abstractions of this representation having a substantial negative effect on heuristic quality. This is also the case for the experiments of the height representation of this domain. The experiments of both representations of the Sliding-Tile Puzzle tell a different story. None of the experimented domain and projection abstractions of either representations of the larger version of this problem domain contain any type 1 spurious transitions. (Type 1 spurious transitions, however, happen in the domain abstractions of the dual representation of the smaller size of this problem domain, 8-puzzle, and are illustrated in Table \ref{tab:SamplePureTransitions}. It may be just coincidental that our choice of abstractions in the larger version of the domain contains none.) The experimented domain abstractions of the stack representation of Towers of Hanoi also contain type 1 spurious transitions with a substantial effect on heuristic quality. This is also true for the tested projection abstractions of the Scanalyzer domain all of which contain type 1 spurious transitions affecting substantially the heuristic quality of their respective abstraction.

Some sample abstractions in different representations of various problem domains containing type 1 spurious transitions are shown in Table \ref{tab:SamplePureTransitions}. The first column shows the problem domain, representation and the size of the corresponding problem domain. The next column illustrates the abstraction rules applied on the given representation of the problem domain described in the first column. The third column shows the number of abstract states, the average heuristic value and average number of expanded nodes by IDA* of the original abstraction. The fourth column shows the same numbers after removing all spurious states and the fifth one shows these numbers after removing both types of spurious transitions. The last column shows the percentage of improvement of average heuristic quality and average number of nodes expanded by IDA* after removing type 1 spurious transitions. The abstractions in this table confirm the fact that type 1 spurious transitions can have a substantial effect on heuristic quality. We should mention again that an abstraction after removing all spurious transitions has the same number of abstract states as the abstraction after removing only spurious states. Type 1 spurious transitions can only decrease heuristic quality.

\begin{center}

\begin{table}[h!]

\begin{footnotesize}

\centerline{\begin{tabular}{|l|l|l|l|l|l|}
\hline 
\rowcolor{LightGrey} {Domain} & {Abstraction} & {Original} & {True} & {Pure} & {\% of} \tabularnewline
\rowcolor{LightGrey} &  &  &  &  & {improvement} \tabularnewline
\rowcolor{LightGrey} &  &  &  &  & {Pure/True} \tabularnewline
\hline 
\rowcolor{LightGrey} {Rep} &  & {Size} & {Size} & {Size} & \tabularnewline \rowcolor{LightGrey} {Size} &  & {Avg.\ $h$} & {Avg.\ $h$} & {Avg.\ $h$} & {Avg.\ $h$} \tabularnewline
\rowcolor{LightGrey} &  & {Avg.\ Nodes} & {Avg.\ Nodes} & {Avg.\ Nodes} & {Avg.\ Nodes} \tabularnewline
\hline
{BW} & {keep [hgh, tp, bln\_on $c,d,e,f$]} & {41,121,810} & {412,800} & {412,800} & \tabularnewline
{height} & {keep [bln\_on\_tp none]} & {13.56} & {18.37} & {19.55} & {6.42} \tabularnewline
{9by3} & keep [hand] & {2,013,949} & {765,404} & {577,189} & {24.59} \tabularnewline
\hline
{STP} & {$2\leftarrow2,8$ } & {181,440} & {181,440} & {181,440} & \tabularnewline
{dual} &  & {13.99} & {13.99} & {21.97} & {57.04}   \tabularnewline
{8-puzzle} &  & {6,584} & {6,584} & {44} & {99.33} \tabularnewline
\hline
{STP} & {$1\leftarrow1,9$ } & {181,440} & {181,440} & {181,440} & \tabularnewline
{dual} &  & {14.95} & {14.95} & {21.97} & {46.96} \tabularnewline
{8-puzzle} &  & {5,357} & {5,357} & {44} & {99.18} \tabularnewline
\hline
{ToH} & {$1\leftarrow1,8$ } & {749,972} & {122,768} & {122,768} & \tabularnewline
{stack} & {$2\leftarrow2,7$ } & {12.77} & {15.46} & {16.94} & {9.57} \tabularnewline
{9by4} & {$3\leftarrow3,6,9$} & {21,553} & {17,317} & {14,971} & {13.55} \tabularnewline
 & {$4\leftarrow4,5$} &  &  &  & \tabularnewline
\hline
{SCNZ} & {keep [belts 4,5,6,7]} & {65,536} & {26,880} & {26,880} & \tabularnewline
{standard} & {keep [bln\_analyzed 4,5,6,7]} & {3.75} & {3.88} & {4.13} & {6.44} \tabularnewline
{8 belts} &  & {1,301,462} & {1,247,700} & {1,157,560} & {7.22} \tabularnewline
\hline
\end{tabular}{\footnotesize }}{\footnotesize \par}

\footnotesize \caption{Sample abstractions containing type 1 spurious transitions. The third column from the right shows the average heuristic value and the average number of nodes expanded by IDA* after removing all spurious states. The second column from the right shows the same measures after removing both types of spurious transitions. The last column shows the percentage of improvement of these measures after removing type 1 spurious transitions  (BW: Blocks World, STP: Sliding-Tile Puzzle, ToH: Towers of Hanoi, SCNZ: Scanalyzer).}
\footnotesize
\par

\label{tab:SamplePureTransitions}

\end{footnotesize}

\end{table}

\par\end{center}

In the extreme situation, it might even be the case that an abstraction not generating any spurious states might contain type 1 spurious transitions. As an example in the dual representation of the 8-puzzle, consider the abstraction that identifies the $2^{nd}$ and $8^{th}$ locations together. As we can see in the second row of Table \ref{tab:SamplePureTransitions}, although the abstraction does not generate any spurious states, type 1 spurious transitions highly affect the quality of the heuristic value, decreasing the average heuristic from 21.97 to 13.99. The abstraction that maps the first and last locations of the puzzle together (row 3) is another example in this representation showing a similar behavior. In this abstraction, type 1 spurious transitions decrease the average heuristic from 21.97 to 14.95.

\subsection{Pattern Database Implementation}\label{sec:pdb}

Pattern databases can be implemented in many different ways and hash tables are among the most popular data structures used for this purpose. Regular hashing, however, has the problem of \emph{address collision} where two (or more) abstract states might be mapped to the same address in the lookup table. This problem is usually resolved using a collision resolution mechanism by using two well-known mechanisms of \emph{open hashing (separate chaining) }and \emph{closed hashing (open addressing)}.

Perfect hash functions for permutations can be used as an alternative approach for storing pattern databases~\citep{MyrvoldR01, DBLP:conf/aaai/KorfS05}. By using these perfect hash functions, no two abstract states are mapped to the same address in the lookup table and therefore collisions are totally avoided. This approach, however, suffers from the problem of unused slots in the hash table corresponding to unreachable abstract states. This can be somewhat justified by the fact that these methods do not store the actual states in the PDB and find the actual states by applying an \emph{unrank} function on the integer hash value of the abstract state. The number of these unused slots depends on the problem domain, representation and corresponding abstraction of the PDB. This approach of storing PDBs can be efficient in some cases and quite inefficient in others. Unless equipped with some domain specific knowledge (like parity\footnote{The parity function divides the state space of the Sliding-Tile Puzzle into two disconnected components; all the states in one component have equal parities (even or odd parity)~\citep{Johnson1879}.} information in the Sliding-Tile Puzzle) to avoid unreachable states, these perfect hash function methods can be quite memory-wasteful. The Constrained Movement Sliding-Tile Puzzle (CSTP) domain is a good example of the inefficiency of this approach for PDB implementation. 

Universal perfect hashing or FKS (Fredman-Koml{\'o}s-Szemer{\'e}di) is a different approach appropriate for implementing PDBs and manages to avoid collisions totally. State-of-the-art minimal perfect hash algorithms using random graphs \citep{conf/wads/BotelhoPZ07} can also be used for PDB implementation\footnote{The fact that they need to know all the keys in advance is not a problem with PDB implementation because a PDB contains an invariant set of abstract states.}. In the meantime, both FKS and minimal perfect hashing schemes need some small space overhead to store hash function representation in memory (for the minimum perfect hash function implemented by BDZ(RAM) algorithm~\citep{conf/wads/BotelhoPZ07}, this space overhead is around $2.5n$ bits for a set of $n$ keys).

Successfully applied to planning and model checking, BDDs are another suitable approach for implementing pattern databases~\citep{DBLP:conf/aips/Edelkamp02, DBLP:conf/aaai/JensenBV02}. Introduced for the purpose of representing pre-computed heuristics, Level-Ordered Edge Sequence (LOES) is another approach for implementing pattern databases in domain-independent planning that claims to be a quite efficient representation of precomputed heuristics~\citep{DBLP:conf/socs/SchmidtZ11}.

\subsection{Note on Pattern Database Size}

As yet, our experimental results discussion about the size of PDBs has only been restricted to the number of states in them. This was done because different well-known methods of implementing PDBs might end up in different sizes in bytes (some of the most popular approaches for PDB implementation are discussed in Section \ref{sec:pdb}). To have a sense of the effect of spurious states on PDB size in bytes, we discuss them in our regular hash table implementation of PDBs and their corresponding sizes in bytes.

In our current PDB implementation, we have used the linear probing variation of open addressing for address collision resolution. Though this approach is prone to primary clustering, due to the cache friendly property of linear probing, it has been shown that linear probing can outperform other hash structures when dealing with load factors of 30\%-70\% \citep{DBLP:conf/alenex/HeilemanL05, DBLP:conf/wae/BlackMQ98}. However, the performance of linear probing is highly reliant on the choice of a good hash function. A good hash function should uniformly distribute keys in the hash table (minimum number of collisions) and should be simple enough to be evaluated fast. It is worth mentioning that although other variations of open addressing are also valid for PDB implementation, we have chosen linear probing for its simplicity of implementation and high cache performance.

In general, if the hash tables become too full ($load\, factor\,\approx\,1$), the performance of all open addressing collision resolution mechanisms degrades a lot. To avoid this problem, we use rehashing when the load factor becomes bigger than 0.75 (considering time-space trade-off, a load factor of 0.75 seems to be a good threshold for rehashing).

\begin{figure}[ht]
\noindent \begin{centering}
\includegraphics[scale=0.34]{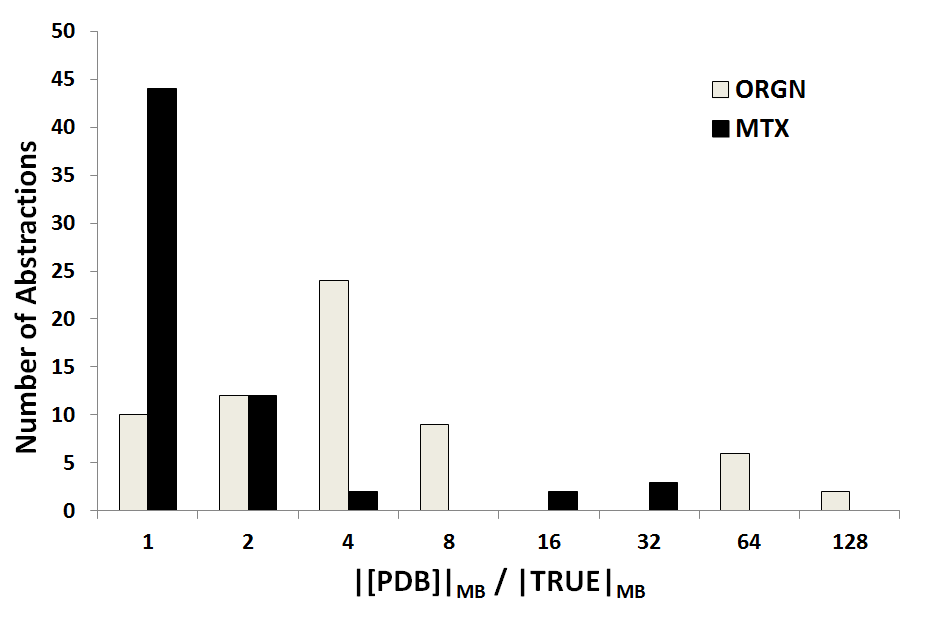}
\par\end{centering}

\caption{Ratio of original PDB size to true PDB size vs.\ the ratio of exhaustive approach mutex-removed PDB size to true PDB size all in unit of megabytes.}

\label{fig:PDBRatioEE-MB} 
\end{figure}

Figure \ref{fig:PDBRatioEE-MB} shows the ratio of \ORGN\ size to \TRUEPDB\ size vs.\ the ratio of \EE\ size to \TRUEPDB\ size in units of megabytes of our experimented abstractions. This figure shows that we can have a major reduction in PDB size after mutex detection (when the relative number of spurious states to the total number of abstract states is significant). It might also be the case that no reduction in PDB size is achieved by mutex detection. This is due to the fact that in our PDB implementation, the hash table sizes are powers of two and we double the table size at 0.75 load factor for performance reasons meaning that the number of entries in the hash table implementation of a PDB is not necessarily equal to the number of states in the abstraction.

In addition to regular hashing, other popular hashing approaches for implementing PDBs like universal perfect hashing, minimal perfect hashing and basically every scalable domain-independent PDB implementation data structure can also benefit from the removal of spurious states. An exception is perfect hashing for permutations\footnote{Notice this is different from PDB representation algorithms that use generic perfect hash functions and minimum perfect hash functions. One such algorithm for efficient representation of PDBs is proposed in \citep{DBLP:conf/aips/SadeqiH16}.}, which does not benefit from the removal of spurious states. However, since it is generally not possible to find such a perfect hash function that is nearly surjective, this approach suffers from the problem of an excessive number of unused slots \citep{DBLP:conf/socs/SchmidtZ11} and therefore is not universally effective for PDB implementation in a domain-independent setting.

\section{Conclusions and Future Work} \label{Section-Conclusions}

We introduced different types of spurious transitions that can be created when using abstraction to define a heuristic. Such transitions can increase the number of abstract states and decrease the quality of the heuristic. We then comprehensively studied the effect of spurious states (type 2 spurious transitions) on heuristic quality and abstraction size on small size problem domains and showed how mutex pair detection can help to neutralize these negative effects to a reasonable extent. Further, we have tested a state-of-the-art mutex pair detection method, $h^2$, on bigger size problem domains to show that mutex pair detection can be effective in improving heuristic quality and decreasing abstraction size in real world problem domains.

We also observed that mutex pair detection can improve the heuristic quality substantially despite the fact that the number of spurious states removed by it is small compared to the number of states in the original PDB. This shows that a small number of spurious states can have a notable effect on the heuristic quality, emphasizing the importance of spurious states detection in general and mutex pair detection in particular. However, from some example abstractions we notice that mutex pair detection is sometimes ineffective in increasing the heuristic values when the mutex-based spurious states are not the ones that have the most damaging effect on the quality of the heuristic. There are even cases where removing all spurious states has a small or no effect on improving heuristic quality illustrating the fact that spurious states do not always have a negative effect on abstraction-based heuristics. Using various example abstractions in small size problem domains, we have also illustrated that transitions not involving spurious states (type 1 spurious transitions) can have a substantial harmful effect on heuristics.

Although mutex pair and spurious state detection can decrease the abstraction size, it is not immediately obvious that the PDB size in bytes is also decreased. For this reason, we have discussed different methods for PDB implementation illustrating the fact that mutex pair and spurious state detection can also decrease PDB size in bytes for most PDB implementation approaches. 

Since our empirical study relies on experiments on some benchmark problem domains, it is important to create or find problem domains with a high ratio of higher order mutexes to mutex pairs to better illustrate the importance of developing more effective methods for detecting spurious transitions in general. This can be done, for example, by adding more constraints to the existing benchmark problem domains or by creating other problem domains from scratch having many harmful higher order mutexes. 

\bibliographystyle{coin}
\bibliography{JournalPaper}

\pagebreak

\appendix

\renewcommand*{\thesection}{\Alph{section}}

\section{} 
\label{Appendix-DomainsAndRepresentations}


In this appendix, we will introduce different problem domains and representations used for the experiments throughout this document. Unless otherwise stated, all experiments and examples discussed in this document are represented using the PSVN notation.



\subsubsection{Domain 1: Towers of Hanoi}
\begin{figure}[ht]
\noindent \begin{centering}
\includegraphics[scale=0.60]{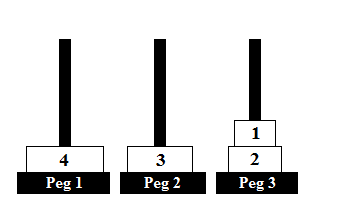}
\par\end{centering}

\caption{A state of Towers of Hanoi with 4 disks and 3 pegs.}

\label{fig:43hanoi-1}
\end{figure}

In the $n$-Disk Towers of Hanoi with $p$ Pegs, a state describes the constellation of $n$ disks stacked on $p$ named pegs. In every move, a disk can be transferred from one peg to another provided that all disks on the destination peg are larger than the moving disk. The goal is to stack up all disks in decreasing order on the goal peg from a given start state using the legal moves. An example state of this domain with 4 disks and 3 pegs is shown in Figure \ref{fig:43hanoi-1}. The following subsections describe the representations used in our experiments.

\subsubsection{Binary Representation}
In the \emph{binary representation}, a state vector has $p \times n$ components, each containing either the values 0 or 1. For every peg a sequence of $n$ components corresponding to the $n$ disks contains the value 1 if that disk is on that peg and 0 if not. For example, the state in Figure \ref{fig:43hanoi-1} would be encoded as $\langle0,0,0,1,0,0,1,0,1,1,0,0\rangle$.

\subsubsection{Disk Representation}
In the \emph{disk representation}, a state vector has $n$ components, one for each disk, containing a value in $\{1,2,...,p\}$ representing the peg on which the disk is located. For example, the state in Figure \ref{fig:43hanoi-1} would be encoded as $\langle 3,3,2,1 \rangle$.

\subsubsection{Stack Representation}

In the \emph{stack representation} of the $n$-Towers of Hanoi with $p$ Pegs, a state is encoded as a vector of length $p(n+1)$, where for every peg a sequence of $n+1$ components encodes the number of disks and the names of disks stacked on this peg (starting from the bottom of the peg); for a stack of $k$ disks, the last $n-k$ components for this peg contain a $0$ where the value $0$ means ``no disk''. For example, the stack representation of the state shown in Figure~\ref{fig:43hanoi-1} would be $\langle1,4,0,0,0,1,3,0,0,0,2,2,1,0,0\rangle$.

\subsubsection{Domain 2: Blocks World with Table Positions}\label{DMN_REP:BW}

\begin{figure}[ht]
\noindent \begin{centering}
\includegraphics[scale=0.60]{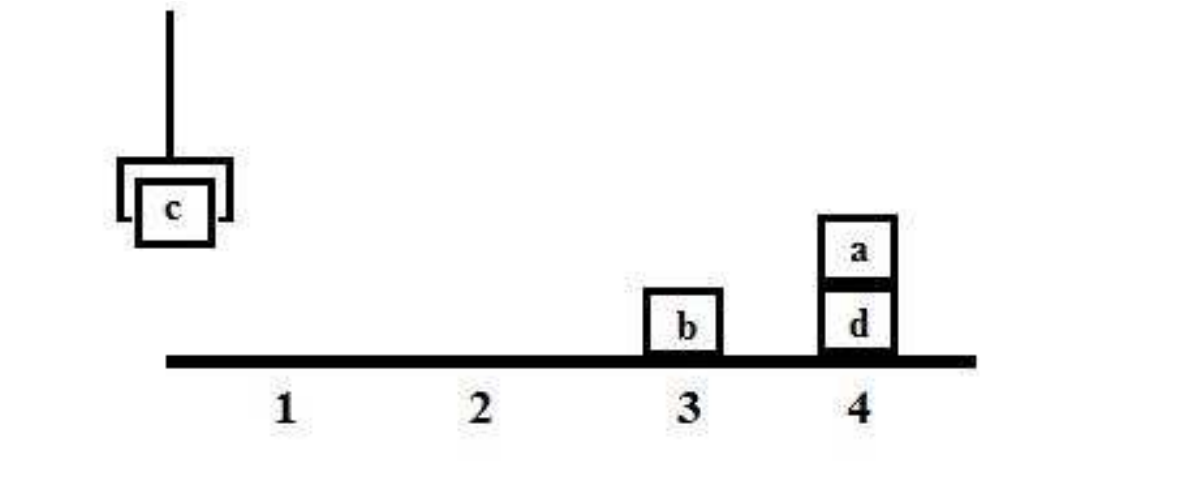}
\par\end{centering}

\caption{A state of the Blocks World with $4$ blocks and $4$ table positions.}

\label{fig:44blocks}
\end{figure}

In the $n$-Blocks World with $p$ Table Positions, a state describes the constellation of $n$ blocks stacked on a table with $p$ named positions, where at most one block can be located in a ``hand''. In every move, either the empty hand picks up the top block off one of the stacks on the table, or the hand holding a block places that block onto an empty table position or on top of a stack of blocks. In all our experiments with this domain, the goal state has all blocks stacked up in increasing lexicographical order, starting with block $a$, on table position 1. Figure \ref{fig:44blocks} shows an example state of this domain with 4 blocks and 4 table positions. The following subsections describe the representations used in our experiments.

\subsubsection{Top Representation}
In the \emph{top representation}, a state vector has $1+p+n$ components each containing either the value $0$ or one of $n$ possible block names:

\begin{itemize}
\item the value of the first component is the name of the block in the hand or 0 if the hand is free,
\item the values of the next $p$ components are the names of the blocks immediately on table positions $1$ through $p$,
\item the values of the last $n$ components are the names of the blocks immediately on top of blocks $a,b,c,\ldots$.
\end{itemize}

where the value 0 means ``no block''. For example, the state in Figure~\ref{fig:44blocks} would be encoded by $\langle c,0,0,b,d,0,0,0,a\rangle$ in the top presentation.

\subsubsection{Height Representation}
In the \emph{height representation}, a state is encoded as a vector of length $1+3n+p$, where

\begin{itemize}
\item the value of the first component is the name of the block in the hand or 0 if the hand is free,
\item for every block a sequence of $3$ components encodes the table position, the height of the block within a stack of blocks (where 1 means that the block is sitting directly on the table), and whether there is any block on top of this block (value 1) or not (value 0).
\item the last $p$ components are flags stating, for each table position, whether there are any blocks on top of it (value 1) or not (value 0).
\end{itemize} 


For example, $\langle c,4,2,0,3,1,0,0,0,0,4,1,1,0,0,1,1\rangle$ is the height representation of the state shown in Figure~\ref{fig:44blocks}.

\subsubsection{Stack Representation}
In the \emph{stack representation}, a state is encoded as a vector of length $1+p(n+1)$, where 


\begin{itemize}

\item the value of the first component is the name of the block in the hand or 0 if the hand is free, 

\item for every table position a sequence of $n+1$ components encodes the number of blocks and the names of blocks stacked on this table position (starting from the bottom of the stack); for a stack of $k$ blocks, the last $n-k$ components for this table position contain a 0. 

\end{itemize}


For example, $\langle c,0,0,0,0,0,0,0,0,0,0,1,b,0,0,0,2,d,a,0,0\rangle$ is the stack representation of the state shown in Figure~\ref{fig:44blocks}. 


\subsubsection{Domain 3: Sliding-Tile Puzzle}
\begin{figure}[ht]
\noindent \begin{centering}
\includegraphics[scale=0.35]{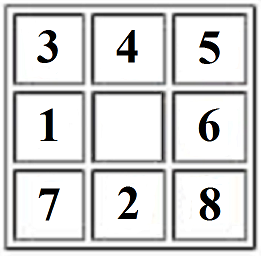}
\par\end{centering}

\caption{A state of the $8$-puzzle.}

\label{fig:8puzzle}
\end{figure}

In the $n\times m$-Sliding-Tile Puzzle\footnote{This problem domain was introduced in Section \ref{Section-Introduction} of the main body of the paper. For the sake of completeness of this section, we have included the problem domain description and representations here again.} \citep{SlocumS06}, representing an $n\times m$ grid, tiles numbered 1 through $n\cdot m-1$ each fill one grid position and the remaining grid position is blank. A move consists of swapping the blank with an adjacent tile. The goal is to have the numbered tiles in increasing order from top left corner to bottom right corner with the blank tile in the bottom right position. Figure \ref{fig:8puzzle} shows an example state of the $3\times3$ Sliding-Tile Puzzle usually known as the $8$-puzzle. The representations used in our experiments are described in the following subsections.

\subsubsection{Standard Representation}
In the \emph{standard representation}, states are vectors of length $n \cdot m$ where each component corresponds to a grid position and represents the number of the tile in this position ($B$, if the position is blank). For example, the state in Figure~\ref{fig:8puzzle} would be represented by $\langle3,4,5,1,B,6,7,2,8\rangle$.

\subsubsection{Dual Representation}
In the \emph{dual representation}, a state is a vector of length $n \cdot m$ where each component corresponds to either one of the numbered tiles or the blank. The value of a vector component represents the grid position at which the corresponding tile is located. For example, the state in Figure~\ref{fig:8puzzle} would be encoded as $\langle4,8,1,2,3,6,7,9,5\rangle$, where the $i^{th}$ component, for $i\le8$, holds the position of tile $i$, and the $9^{th}$ component holds the position of the blank.

\subsubsection{Domain 4: Scanalyzer}\label{DMN_REP:SCNZ}

\begin{figure}[ht]
\noindent \begin{centering}
\includegraphics[scale=0.3]{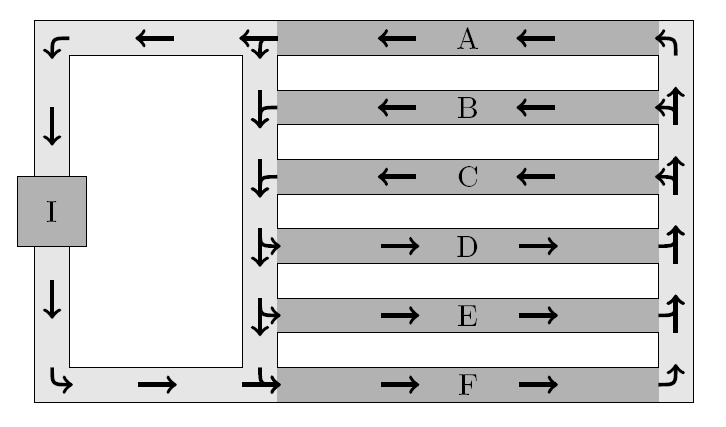}
\par\end{centering}

\caption{6-Belt Scanalyzer, adopted from \citep{DBLP:conf/aips/HelmertL10}. Arrows indicate legal moves.}

\label{fig:scanalyzer}
\end{figure}

In the $n$-Belt Scanalyzer domain, a state describes the placement of $n$ plant batches on $n$ conveyor belts along with information indicating which batches have been ``analyzed'' (for a detailed description of this domain, see \citep{DBLP:conf/aips/HelmertL10}). In a \emph{rotate\/} move, a batch can be switched from one conveyor belt in the upper half (A, B and C in Figure \ref{fig:scanalyzer}) to one in the lower half (D, E and F in Figure \ref{fig:scanalyzer}) and vice versa. In a \emph{rotate-and-analyze}\/ move, a batch can simultaneously be transferred and analyzed from the topmost conveyor belt to the bottommost one while the batch at the bottommost conveyor belt is moved to the topmost one without any change to its ``analyzed'' state. Once a batch is analyzed, it will remain analyzed henceforward.

\subsubsection{Standard Representation}
In the \emph{standard representation} of the $n$-Belt Scanalyzer, a state is encoded as a vector of length $2n$ in which each conveyor belt corresponds to two components: the name of the batch on that belt and a flag indicating whether that batch is analyzed or not. The goal state corresponds to having all plant batches analyzed and replaced back on their original conveyor belts.

\subsubsection{Domain 5: Constrained-Movement Sliding-Tile Puzzle} \label{subsec:CMSTP}

\begin{figure}[ht]
\noindent \begin{centering}
\includegraphics[scale=0.25]{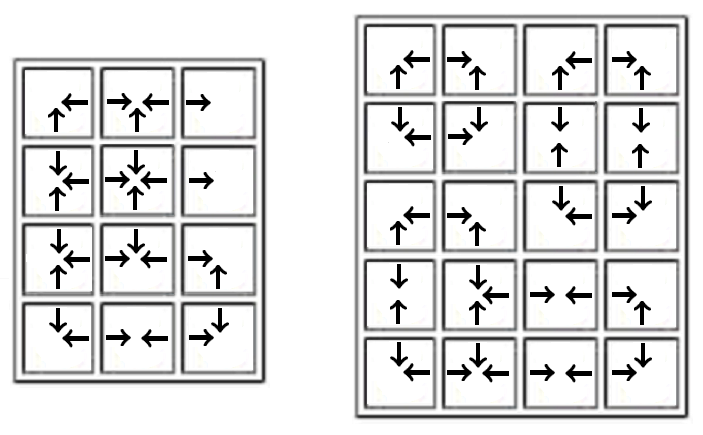}
\par\end{centering}

\caption{\label{fig:CMSP}$3\times4$ and $4\times5$ Constrained-Movement
Sliding-Tile Puzzle. The arrows indicate the possible movements of the tiles, based on the blank location.}

\end{figure}

We modified the Sliding-Tile Puzzle by disallowing some of the tile movements. Two versions of this ``Constrained-Movement Sliding-Tile Puzzle'' were used---see Figure \ref{fig:CMSP}---encoded in the same way as the standard representation of the original Sliding-Tile Puzzle. Note that, in both constrained versions, all the operators are deliberately made invertible.

The Constrained-Movement Sliding-Tile Puzzle domain was designed with the purpose of creating abstractions whose most harmful spurious states contain higher-order abstraction-based mutexes but no abstraction-based mutex pairs. For example, in the $3\times4$ version
of this puzzle, the domain abstraction mapping tile 11 to a blank results in 239,500,800 abstract states, 237,827,520 of which are spurious. While 236,960,640 of these spurious states contain abstraction-based mutex pairs, their removal does not change the resulting heuristic value, averaged over the entire original state space. Removing the remaining 866,880 spurious states, which contain only higher-order abstraction-based mutexes, increases the average heuristic value from 31.94 to 32.75. We observed a similar phenomenon in our experiments with the top representation of Blocks World.

\subsubsection{Domain 6: Depot}
The goal in this problem domain is to have crates stacked in their proper order at their target depots. The actions are to move crates from one depot to another one using trucks, to stack or unstack crates on a fixed number of pallets at every depot and to load or unload trucks with crates using hoists available at fixed locations. The order of crates in the truck does not matter. This domain can be considered as a combination of the Simple Logistics and Blocks World domains, having an element of transportation combined with a Blocks World element. The transportation element of the task is to move crates from one depot to another using trucks. The Blocks World element is to stack and unstack crates and is different from the standard Blocks World in that there are multiple arms called hoists and there are a limited number of named table positions called pallets~\citep{DBLP:books/sp/Helmert2008}. Another difference with the standard Blocks World is that, here, hoists are also used for loading or unloading crates to or from trucks. We used problems from this domain that were also used in planning competitions. For our experiments, the PSVN representation corresponding to the multi-valued representations derived by Fast Downward's preprocessing algorithm is used.


\subsubsection{Domain 7: Storage}
In the Storage planning domain, the task is to move a given number of crates from some containers to some depots using hoists. A specified spatial map connecting different areas of the depot determines the hoist movement inside a depot meaning that this planning domain requires spatial reasoning. Different numbers of depots, hoists, crates, containers, and the spatial map of the depot areas defines a problem instance for this planning domain. We used problems from this domain that were also used in planning competitions. For our experiments, the PSVN representation corresponding to the multi-valued representations derived by Fast Downward's preprocessing algorithm is used.

\subsection{Details of the Abstractions}\label{AppendixB-1}

In this subsection we present the details of the abstractions discussed in Section \ref{sec:MutexDetection} of the main body of the paper. For a problem domain in a given representation, each table gives a brief description of how each abstraction was defined along with different measures for three different PDBs. The first column illustrates the abstraction rules applied. The next three columns respectively report the results for the original PDB, the PDB after removing all mutex pairs and the PDB after removing all spurious states. For each PDB, the abstraction size in terms of the number of abstract states and memory size, the average heuristic value and the average number of nodes expanded using IDA* are calculated and shown in columns 2, 3 and 4. The heuristic value of the entire original state space is averaged to obtain the average heuristic value of each PDB. To calculate the average number of nodes expanded using IDA*, we first sample 1,000 start states uniformly at random, solve these problem instances using each PDB and average the number of nodes expanded.

\setlength\extrarowheight{2pt}

\begin{footnotesize}

\begin{longtable}{|l|l|l|l|}

\hline
\rowcolor{LightGrey} Abstraction  & Original  & Mutex  & True\tabularnewline
\hline
\rowcolor{LightGrey}  & Size  & Size  & Size\tabularnewline
\rowcolor{LightGrey}  & Avg.\ $h$  & Avg.\ $h$  & Avg.\ $h$\tabularnewline
\rowcolor{LightGrey}  & Avg.\ Nodes  & Avg.\ Nodes  & Avg.\ Nodes\tabularnewline
\hline
\endfirsthead
\multicolumn{4}{l}%
{\tablename\ \thetable\ -- \textit{Continued from previous page}} \\
\hline
\rowcolor{LightGrey} Abstraction  & Original  & Mutex  & True\tabularnewline
\hline
\rowcolor{LightGrey}  & Size  & Size  & Size\tabularnewline
\rowcolor{LightGrey}  & Avg.\ $h$  & Avg.\ $h$  & Avg.\ $h$\tabularnewline
\rowcolor{LightGrey}  & Avg.\ Nodes  & Avg.\ Nodes  & Avg.\ Nodes\tabularnewline
\hline
\endhead

$a\leftarrow a,b,c$  & 974,364 (41 MB)  & 974,364 (41 MB)  & 951,990 (41 MB)\tabularnewline
$b\leftarrow d,e,f$  & 19.22  & 19.22  & 21.06\tabularnewline
$c\leftarrow g,h,i$  & 553,951  & 553,951  & 350,209\tabularnewline
\hline 
$a\leftarrow a,b,c,d$  & 732,504 (21 MB)  & 724,280 (21 MB)  & 712,252 (21 MB)\tabularnewline
$b\leftarrow e,f,g$  & 18.93  & 20.82  & 21.49\tabularnewline
$c\leftarrow h,i$  & 588,698  & 380,241  & 314,228\tabularnewline
\hline 
$a\leftarrow a,b,c,d$  & 732,504 (21 MB)  & 724,280 (21 MB)  & 712,252 (21 MB)\tabularnewline
$c\leftarrow g,h,i$  & 18.68  & 19.52  & 20.88\tabularnewline
$b\leftarrow e,f$  & 620,015  & 518,211  & 370,248\tabularnewline
\hline 
$b\leftarrow c,d,e,f$  & 732,504 (21 MB)  & 724,280 (21 MB)  & 712,252 (21 MB)\tabularnewline
$c\leftarrow g,h,i$  & 18.63  & 18.99  & 20.60\tabularnewline
$a\leftarrow a,b$  & 625,933  & 579,070  & 394,280\tabularnewline
\hline 
$c\leftarrow e,f,g,h,i$  & 440,280 (21 MB)  & 430,540 (21 MB)  & 425,240 (21 MB)\tabularnewline
$a\leftarrow a,b$  & 17.40  & 18.35  & 19.02\tabularnewline
$b\leftarrow c,d$  & 788,965  & 657,698  & 570,176\tabularnewline
\hline 
$b\leftarrow b,c,d,h,i$  & 83,400 (2.6 MB)  & 83,400 (2.6 MB)  & 82,445 (2.6 MB)\tabularnewline
$a\leftarrow a,e,f,g$  & 12.72  & 12.72{*}  & 12.72\tabularnewline
 & 1,541,416  & 1,541,416  & 1,541,404\tabularnewline
\hline 
$b\leftarrow d,e,f,g,h,i$  & 55,365 (2.6 MB)  & 55,365 (2.6 MB)  & 54,591 (2.6 MB)\tabularnewline
$a\leftarrow a,b,c$  & 13.01  & 13.01  & 13.71\tabularnewline
 & 1,493,788  & 1,493,788  & 1,370,657\tabularnewline
\hline 
\multicolumn{1}{c}{} & \multicolumn{1}{c}{} & \multicolumn{1}{c}{} & \multicolumn{1}{c}{} \tabularnewline
\multicolumn{1}{c}{} & \multicolumn{1}{c}{} & \multicolumn{1}{c}{} & \multicolumn{1}{c}{} \tabularnewline
\multicolumn{1}{c}{} & \multicolumn{1}{c}{} & \multicolumn{1}{c}{} & \multicolumn{1}{c}{} \tabularnewline
\multicolumn{1}{c}{} & \multicolumn{1}{c}{} & \multicolumn{1}{c}{} & \multicolumn{1}{c}{} \tabularnewline
$b\leftarrow b,c,d,e,f,g,h$  & 23,541 (641 KB)  & 23,303 (641 KB)  & 23,037 (641 KB)\tabularnewline
$a\leftarrow a,i$  & 11.62  & 11.62  & 11.67\tabularnewline
 & 1,730,676  & 1,730,676  & 1,722,200\tabularnewline
\hline

\caption{Domain abstractions of the 9-Blocks World with 3 Table Positions in the top representation. This state space has 36,288,000 reachable original states with an average distance of $37.11$.}\label{tab:BW-TOP-9by3-CA}

\end{longtable}

\end{footnotesize}

\begin{footnotesize}

\begin{longtable}{|l|l|l|l|}

\hline
\rowcolor{LightGrey} Abstraction  & Original  & Mutex  & True\tabularnewline
\hline
\rowcolor{LightGrey} & Size  & Size  & Size\tabularnewline
\rowcolor{LightGrey} & Avg.\ $h$  & Avg.\ $h$  & Avg.\ $h$\tabularnewline
\rowcolor{LightGrey} & Avg.\ Nodes  & Avg.\ Nodes  & Avg.\ Nodes\tabularnewline
\hline
\endfirsthead
\multicolumn{4}{l}%
{\tablename\ \thetable\ -- \textit{Continued from previous page}} \\
\hline
\rowcolor{LightGrey} Abstraction  & Original  & Mutex  & True\tabularnewline
\hline
\rowcolor{LightGrey} & Size  & Size  & Size\tabularnewline
\rowcolor{LightGrey} & Avg.\ $h$  & Avg.\ $h$  & Avg.\ $h$\tabularnewline
\rowcolor{LightGrey} & Avg.\ Nodes  & Avg.\ Nodes  & Avg.\ Nodes\tabularnewline
\hline
\endhead

keep [hgh, tp, bln\_on $c,d,e,f$]  & 41,121,810 (2.3 GB)  & 22,587,552 (1.2 GB)  & 412,800 (37 MB)\tabularnewline
keep [bln\_on\_tp none]  & 13.56  & 13.67  & 18.37\tabularnewline
keep [hand] & 2,013,949  & 1,975,367  & 765,404\tabularnewline
\hline 
keep [hgh, tp, bln\_on $c,f,g,h${]}  & 41,121,810 (2.3 GB)  & 22,587,552 (1.2 GB)  & 412,800 (37 MB)\tabularnewline
keep [bln\_on\_tp none]  & 14.06  & 14.18  & 20.43\tabularnewline
keep [hand] & 1,840,035  & 1,803,623  & 465,554\tabularnewline
\hline 
keep [hgh, tp, bln\_on $b,d,f,h$]  & 41,121,810 (2.3 GB)  & 22,587,552 (1.2 GB)  & 412,800 (37 MB)\tabularnewline
keep [bln\_on\_tp none]  & 15.90  & 15.96  & 23.22\tabularnewline
keep [hand] & 1,279,181  & 1,265,986  & 230,817\tabularnewline
\hline 
keep [hgh, tp, bln\_on $c,f,g$]  & 7,358,229 (577 MB)  & 1,278,048 (73 MB)  & 94,320 (4.6 MB)\tabularnewline
keep [bln\_on\_tp all] & 12.88  & 13.58  & 19.23\tabularnewline
keep [hand] & 2,236,377  & 1,969,702  & 630,115\tabularnewline
\hline 
keep [hgh, tp, bln\_on $e,f,h$]  & 7,358,229 (577 MB)  & 1,278,048 (73 MB)  & 94,320 (4.6 MB)\tabularnewline
keep [bln\_on\_tp all] & 12.98  & 13.75  & 17.52\tabularnewline
keep [hand] & 2,195,248  & 1,908,391  & 924,221\tabularnewline
\hline 
keep [tp $a,b,c,d,e,f,g,h,i${]}  & 576,441 (37 MB)  & 78,732 (4.6 MB)  & 78,732 (4.6 MB)\tabularnewline
keep [bln\_on\_tp all] & 11.85  & 11.85  & 11.85\tabularnewline
keep [hand] & 2,512,162  & 2,512,162  & 2,512,162\tabularnewline
\hline 
keep [hgh, tp, bln\_on $a,b${]}  & 164,679 (9.1 MB)  & 46,464 (2.3 MB)  & 12,810 (1.2 MB)\tabularnewline
keep [bln\_on\_tp all] & 10.83  & 11.55  & 13.49\tabularnewline
keep [hand] & 3,058,623  & 2,721,369  & 2,006,458\tabularnewline
\hline 
keep [hgh, tp, bln\_on $h,i$]  & 164,679 (9.1 MB)  & 46,464 (2.3 MB)  & 12,810 (1.2 MB)\tabularnewline
keep [bln\_on\_tp all] & 9.52  & 10.76  & 13.43\tabularnewline
keep [hand] & 3,709,104  & 3,089,660  & 2,011,106\tabularnewline
\hline

\caption{Projection abstractions of the 9-Blocks World with 3 Table Positions in the height representation. This state space has 36,288,000 reachable original states with an average distance of $37.11$.}\label{tab:BW-POS_HEIGHT-9by3}

\end{longtable}


\end{footnotesize}

\begin{footnotesize}

\begin{longtable}{|l|l|l|l|}

\hline
\rowcolor{LightGrey} Abstraction  & Original  & Mutex  & True\tabularnewline
\hline
\rowcolor{LightGrey}  & Size  & Size  & Size\tabularnewline
\rowcolor{LightGrey}  & Avg.\ $h$  & Avg.\ $h$  & Avg.\ $h$\tabularnewline
\rowcolor{LightGrey}  & Avg.\ Nodes  & Avg.\ Nodes  & Avg.\ Nodes\tabularnewline
\hline
\endfirsthead
\multicolumn{4}{l}%
{\tablename\ \thetable\ -- \textit{Continued from previous page}} \\
\hline
\rowcolor{LightGrey} Abstraction  & Original  & Mutex  & True\tabularnewline
\hline
\rowcolor{LightGrey}  & Size  & Size  & Size\tabularnewline
\rowcolor{LightGrey}  & Avg.\ $h$  & Avg.\ $h$  & Avg.\ $h$\tabularnewline
\rowcolor{LightGrey}  & Avg.\ Nodes  & Avg.\ Nodes  & Avg.\ Nodes\tabularnewline
\hline
\endhead

keep [locations 4,5,6,7,8,9,10]  & 35,831,808 (1.1 GB)  & 3,991,680 (129 MB)  & 3,991,680 (129 MB)\tabularnewline
 & 11.99  & 12.39  & 12.39\tabularnewline
 & 43,492,593  & 36,175,632  & 36,175,632 \tabularnewline
\hline 
\multicolumn{1}{c}{} & \multicolumn{1}{c}{} & \multicolumn{1}{c}{} & \multicolumn{1}{c}{} \tabularnewline
keep [locations 3,4,6,8,10,11,12]  & 35,831,808 (1.1 GB)  & 3,991,680 (129 MB)  & 3,991,680 (129 MB)\tabularnewline
 & 12.20  & 13.54  & 13.54\tabularnewline
 & 37,560,310  & 21,126,171  & 21,126,171 \tabularnewline
\hline 
keep [locations 5,6,7,8,9,10]  & 2,985,984 (65 MB)  & 665,280 (17 MB)  & 665,280 (17 MB)\tabularnewline
 & 9.54  & 9.78  & 9.78\tabularnewline
 & 126,982,723  & 114,691,425  & 114,691,425 \tabularnewline
\hline 
keep [locations 3,4,5,8,9,11]  & 2,985,984 (65 MB)  & 665,280 (17 MB)  & 665,280 (17 MB)\tabularnewline
 & 9.86  & 10.22  & 10.22\tabularnewline
 & 109,545,320  & 93,658,444  & 93,658,444 \tabularnewline
\hline 
keep [locations 1,4,5,8,9,11]  & 2,985,984 (65 MB)  & 665,280 (17 MB)  & 665,280 (17 MB)\tabularnewline
 & 9.70  & 9.99  & 9.99\tabularnewline
 & 117,457,170  & 103,145,073  & 103,145,073 \tabularnewline
\hline
keep [locations 6,7,8,9,12]  & 248,832 (8.1 MB)  & 95,040 (2.1 MB)  & 95,040 (2.1 MB)\tabularnewline
 & 8.59  & 9.14  & 9.14\tabularnewline
 & 196,542,201  & 154,047,200  & 154,047,200 \tabularnewline
\hline 
keep [locations 6,7,8,9,10]  & 248,832 (8.1 MB)  & 95,040 (2.1 MB)  & 95,040 (2.1 MB)\tabularnewline
 & 8.11  & 8.28  & 8.28\tabularnewline
 & 239,324,271  & 222,887,668  & 222,887,668 \tabularnewline
\hline 

\caption{Projection abstractions of the $3 \times 4$ Sliding-Tile Puzzle in the standard representation. This state space has 239,500,800 reachable original states with an average distance of 35.10.}\label{tab:SP-ST-3by4-Continued-1}

\end{longtable} 

\end{footnotesize}

\begin{footnotesize}

\begin{longtable}{|l|l|l|l|}

\hline
\rowcolor{LightGrey} Abstraction  & Original  & Mutex  & True\tabularnewline
\hline
\rowcolor{LightGrey}  & Size  & Size  & Size\tabularnewline
\rowcolor{LightGrey}  & Avg.\ $h$  & Avg.\ $h$  & Avg.\ $h$\tabularnewline
\rowcolor{LightGrey}  & Avg.\ Nodes  & Avg.\ Nodes  & Avg.\ Nodes\tabularnewline
\hline
\endfirsthead
\multicolumn{4}{l}%
{\tablename\ \thetable\ -- \textit{Continued from previous page}} \\
\hline
\rowcolor{LightGrey} Abstraction  & Original  & Mutex  & True\tabularnewline
\hline
\rowcolor{LightGrey}  & Size  & Size  & Size\tabularnewline
\rowcolor{LightGrey}  & Avg.\ $h$  & Avg.\ $h$  & Avg.\ $h$\tabularnewline
\rowcolor{LightGrey}  & Avg.\ Nodes  & Avg.\ Nodes  & Avg.\ Nodes\tabularnewline
\hline
\endhead

keep [tiles 1,2,7,8,9,10,11]  & 35,831,808 (1.1 GB)  & 3,991,680 (129 MB)  & 3,991,680 (129 MB)\tabularnewline
 & 15.00  & 16.57  & 16.57\tabularnewline
 & 11,691,317  & 5,749,346  & 5,749,346\tabularnewline
\hline 
keep [tiles 1,2,5,6,9,10,blank]  & 35,831,808 (1.1 GB)  & 3,991,680 (129 MB)  & 3,991,680 (129 MB)\tabularnewline
 & 23.65  & 27.83  & 27.83\tabularnewline
 & 395,325  & 74,476  & 74,476\tabularnewline
\hline 
keep [tiles 5,6,7,8,9,10]  & 2,985,984 (65 MB)  & 665,280 (17 MB)  & 665,280 (17 MB)\tabularnewline
 & 11.83  & 13.11  & 13.11\tabularnewline
 & 49,239,277  & 27,878,480  & 27,878,480\tabularnewline
\hline 
keep [tiles 1,4,5,6,10,11]  & 2,985,984 (65 MB)  & 665,280 (17 MB)  & 665,280 (17 MB)\tabularnewline
 & 12.83  & 13.70  & 13.70\tabularnewline
 & 31,869,485  & 21,398,011  & 21,398,011\tabularnewline
\hline 
keep [tiles 1,2,5,6,9,11]  & 2,985,984 (65 MB)  & 665,280 (17 MB)  & 665,280 (17 MB)\tabularnewline
 & 12.50  & 13.43  & 13.43\tabularnewline
 & 36,136,853  & 23,629,959  & 23,629,959\tabularnewline
\hline 
keep [tiles 1,3,5,7,9,11]  & 2,985,984 (65 MB)  & 665,280 (17 MB)  & 665,280 (17 MB)\tabularnewline
 & 12.83  & 13.52  & 13.52\tabularnewline
 & 32,045,751  & 23,320,963  & 23,320,963\tabularnewline
\hline 
keep [tiles 6,7,8,9,blank]  & 248,832 (8.1 MB)  & 95,040 (2.1 MB)  & 95,040 (2.1 MB)\tabularnewline
 & 16.40  & 20.33  & 20.33\tabularnewline
 & 7,767,948  & 1,774,690  & 1,774,690\tabularnewline
\hline 
\multicolumn{1}{c}{} & \multicolumn{1}{c}{} & \multicolumn{1}{c}{} & \multicolumn{1}{c}{} \tabularnewline
\multicolumn{1}{c}{} & \multicolumn{1}{c}{} & \multicolumn{1}{c}{} & \multicolumn{1}{c}{} \tabularnewline
keep [tiles 3,4,5,9,blank]  & 248,832 (8.1 MB)  & 95,040 (2.1 MB)  & 95,040 (2.1 MB)\tabularnewline
 & 18.04  & 21.10  & 21.10\tabularnewline
 & 4,097,139  & 1,367,507  & 1,367,507\tabularnewline
\hline 
keep [tiles 4,5,8,11,blank]  & 248,832 (8.1 MB)  & 95,040 (2.1 MB)  & 95,040 (2.1 MB)\tabularnewline
 & 16.14  & 19.60  & 19.60\tabularnewline
 & 8,544,712  & 2,279,383  & 2,279,383\tabularnewline
\hline 
keep [tiles 2,4,7,11,blank]  & 248,832 (8.1 MB)  & 95,040 (2.1 MB)  & 95,040 (2.1 MB)\tabularnewline
 & 18.19  & 21.39  & 21.39\tabularnewline
 & 3,710,707  & 1,226,853  & 1,226,853\tabularnewline
\hline 
keep [tiles 2,3,5,10,blank]  & 248,832 (8.1 MB)  & 95,040 (2.1 MB)  & 95,040 (2.1 MB)\tabularnewline
 & 19.99  & 23.16  & 23.16\tabularnewline
 & 1,815,783  & 635,318  & 635,318\tabularnewline
\hline 
keep [tiles 1,4,7,10,blank]  & 248,832 (8.1 MB)  & 95,040 (2.1 MB)  & 95,040 (2.1 MB)\tabularnewline
 & 19.93  & 23.50  & 23.50\tabularnewline
 & 1,880,249  & 533,181  & 533,181\tabularnewline
\hline 
keep [tiles 2,3,6,8,blank]  & 248,832 (8.1 MB)  & 95,040 (2.1 MB)  & 95,040 (2.1 MB)\tabularnewline
 & 18.20  & 21.86  & 21.86\tabularnewline
 & 3,565,727  & 980,612  & 980,612\tabularnewline
\hline 
keep [tiles 1,4,5,11,blank]  & 248,832 (8.1 MB)  & 95,040 (2.1 MB)  & 95,040 (2.1 MB)\tabularnewline
 & 18.47  & 21.56  & 21.56\tabularnewline
 & 3,426,608  & 1,169,477  & 1,169,477\tabularnewline
\hline 
keep [tiles 3,4,8,9,blank]  & 248,832 (8.1 MB)  & 95,040 (2.1 MB)  & 95,040 (2.1 MB)\tabularnewline
 & 17.98  & 21.13  & 21.13\tabularnewline
 & 4,135,465  & 1,361,603  & 1,361,603\tabularnewline
\hline 
keep [tiles 1,7,8,9]  & 20,736 (513 KB)  & 11,880 (257 KB)  & 11,880 (257 KB)\tabularnewline
 & 8.17  & 8.43  & 8.43\tabularnewline
 & 251,115,269  & 222,500,056  & 222,500,056\tabularnewline
\hline

\caption{Projection abstractions of the $3 \times 4$ Sliding-Tile Puzzle in the dual representation. This state space has 239,500,800 reachable original states with an average distance of 35.10.}\label{tab:SP-DU-3by4-1}

\end{longtable} 

\end{footnotesize}

\begin{footnotesize}

\begin{longtable}{|l|l|l|l|}

\hline
\rowcolor{LightGrey} Abstraction  & Original  & Mutex  & True\tabularnewline
\hline
\rowcolor{LightGrey}  & Size  & Size  & Size\tabularnewline
\rowcolor{LightGrey}  & Avg.\ $h$  & Avg.\ $h$  & Avg.\ $h$\tabularnewline
\rowcolor{LightGrey}  & Avg.\ Nodes  & Avg.\ Nodes  & Avg.\ Nodes\tabularnewline
\hline
\endfirsthead
\multicolumn{4}{l}%
{\tablename\ \thetable\ -- \textit{Continued from previous page}} \\
\hline
\rowcolor{LightGrey} Abstraction  & Original  & Mutex  & True\tabularnewline
\hline
\rowcolor{LightGrey}  & Size  & Size  & Size\tabularnewline
\rowcolor{LightGrey}  & Avg.\ $h$  & Avg.\ $h$  & Avg.\ $h$\tabularnewline
\rowcolor{LightGrey}  & Avg.\ Nodes  & Avg.\ Nodes  & Avg.\ Nodes\tabularnewline
\hline
\endhead

$2\leftarrow2,6$  & 977,808 (89 MB)  & 507,832 (45 MB)  & 204,783 (204,784) (23 MB)\tabularnewline
$3\leftarrow3,7$  & 14.84  & 15.36  & 15.90\tabularnewline
$4\leftarrow4,8$  & 18,306  & 17,494  & 16,644\tabularnewline
$5\leftarrow5,9$  &  &  & \tabularnewline
\hline 
$1\leftarrow1,2,6$  & 878,092 (89 MB)  & 358,544 (23 MB)  & 123,188 (12 MB)\tabularnewline
$3\leftarrow3,7$  & 13.54  & 14.11  & 14.83\tabularnewline
$4\leftarrow4,8$  & 20,342  & 19,446  & 18,313\tabularnewline
$5\leftarrow5,9$  &  &  & \tabularnewline
\hline 
$1\leftarrow1,9$  & 850,052 (89 MB)  & 434,560 (45 MB)  & 224,512 (23 MB)\tabularnewline
$2\leftarrow2,6$  & 16.60  & 18.57  & 23.85 \tabularnewline
$5\leftarrow5,8$  & 15,534  & 12,451  & 4,341\tabularnewline
\hline 
\multicolumn{1}{c}{} & \multicolumn{1}{c}{} & \multicolumn{1}{c}{} & \multicolumn{1}{c}{} \tabularnewline
\multicolumn{1}{c}{} & \multicolumn{1}{c}{} & \multicolumn{1}{c}{} & \multicolumn{1}{c}{} \tabularnewline
\multicolumn{1}{c}{} & \multicolumn{1}{c}{} & \multicolumn{1}{c}{} & \multicolumn{1}{c}{} \tabularnewline
$1\leftarrow1,8$  & 749,972 (45 MB)  & 505,148 (45 MB)  & 122,768 (12 MB)\tabularnewline
$2\leftarrow2,7$  & 12.77  & 13.07  & 15.46\tabularnewline
$3\leftarrow3,6,9$  & 21,553  & 21,085  & 17,317\tabularnewline
$4\leftarrow4,5$  &  &  & \tabularnewline
\hline 
$1\leftarrow1,7,8,9$  & 639,216 (45 MB)  & 242,520 (23 MB)  & 80,016 (5.6 MB)\tabularnewline
 & 17.39  & 19.18  & 20.91\tabularnewline
 & 14,294  & 11,522  & 9,002\tabularnewline
\hline 
$1\leftarrow1,3,5,9$  & 327,004 (23 MB)  & 206,000 (23 MB)  & 150,296 (12 MB)\tabularnewline
 & 17.86  & 19.81  & 20.21\tabularnewline
 & 13,527  & 10,456  & 9,822\tabularnewline
\hline 
$3\leftarrow3,5,7,9$  & 295,632 (23 MB)  & 191,696 (12 MB)  & 116,144 (12 MB)\tabularnewline
$2\leftarrow2,8$  & 14.69  & 15.37  & 16.90\tabularnewline
 & 18,540  & 17,461  & 15,061\tabularnewline
\hline 
$1\leftarrow1,5,6$  & 244,108 (23 MB)  & 154,484 (12 MB)  & 53,516 (5.6 MB)\tabularnewline
$2\leftarrow2,3,8,9$  & 13.15  & 13.30  & 15.40\tabularnewline
 & 20,960  & 20,724  & 17,398\tabularnewline
\hline 
$2\leftarrow2,5$  & 229,264 (23 MB)  & 178,000 (23 MB)  & 148,480 (12 MB)\tabularnewline
$8\leftarrow8,9$  & 20.20  & 22.49  & 23.18\tabularnewline
 & 9,870  & 6,420  & 5,524\tabularnewline
\hline 
$1\leftarrow1,3$  & 228,056 (23 MB)  & 117,504 (12 MB)  & 80,072 (5.6 MB)\tabularnewline
$2\leftarrow2,5,7,9$  & 13.91  & 14.15  & 14.69\tabularnewline
$4\leftarrow4,6$  & 19,758  & 19,373  & 18,515\tabularnewline
$5\leftarrow8$  &  &  & \tabularnewline
\hline 
$1\leftarrow1,4,5,9$  & 81,788 (5.6 MB)  & 46,988 (2.8 MB)  & 28,336 (2.8 MB)\tabularnewline
$3\leftarrow3,6,7,8$  & 11.43  & 11.46  & 11.67\tabularnewline
 & 23,667  & 23,614  & 23,290\tabularnewline
\hline 
$1\leftarrow1,2$  & 8,200 (705 KB)  & 6,800 (705 KB)  & 6,800 (705 KB)\tabularnewline
$3\leftarrow3,4,6,7,8,9$  & 11.17  & 11.17  & 11.17\tabularnewline
 & 24,074  & 24,074  & 24,074\tabularnewline
\hline

\caption{Domain abstractions of the 9-Disks Tower of Hanoi with 4 Pegs in the stack representation. This state space has 262,144 reachable original states with an average distance of $29.39$.}\label{tab:TH-PS-9by4}

\end{longtable} 

\end{footnotesize}

\begin{footnotesize}

\begin{longtable}{|l|l|l|l|}

\hline
\rowcolor{LightGrey} Abstraction  & Original  & Mutex  & True\tabularnewline
\hline
\rowcolor{LightGrey}  & Size  & Size  & Size\tabularnewline
\rowcolor{LightGrey}  & Avg.\ $h$  & Avg.\ $h$  & Avg.\ $h$\tabularnewline
\rowcolor{LightGrey}  & Avg.\ Nodes  & Avg.\ Nodes  & Avg.\ Nodes\tabularnewline
\hline
\endfirsthead
\multicolumn{4}{l}%
{\tablename\ \thetable\ -- \textit{Continued from previous page}} \\
\hline
\rowcolor{LightGrey} Abstraction  & Original  & Mutex  & True\tabularnewline
\hline
\rowcolor{LightGrey}  & Size  & Size  & Size\tabularnewline
\rowcolor{LightGrey}  & Avg.\ $h$  & Avg.\ $h$  & Avg.\ $h$\tabularnewline
\rowcolor{LightGrey}  & Avg.\ Nodes  & Avg.\ Nodes  & Avg.\ Nodes\tabularnewline
\hline
\endhead

keep [belts 3,4,5]  & 13,824 (513 KB)  & 7,680 (257 KB)  & 7,680 (257 KB)\tabularnewline
keep [bln\_analyzed all] & 5.86  & 5.92  & 5.92 \tabularnewline
 & 775  & 754  & 754 \tabularnewline
\hline 
keep [belts 0,1,2]  & 13,824 (513 KB)  & 7,680 (257 KB)  & 7,680 (257 KB)\tabularnewline
keep [bln\_analyzed all] & 5.95  & 5.99  & 5.99 \tabularnewline
 & 737  & 722  & 722\tabularnewline
\hline 
keep [belts 1,3,5] & 13,824 (513 KB)  & 7,680 (257 KB)  & 7,680 (257 KB)\tabularnewline
keep [bln\_analyzed all] & 5.93  & 6.05  & 6.05\tabularnewline
 & 747  & 699  & 699\tabularnewline
\hline 
keep [belts 0,2,4]  & 13,824 (513 KB)  & 7,680 (257 KB)  & 7,680 (257 KB)\tabularnewline
keep [bln\_analyzed all] & 6.04  & 6.12  & 6.12\tabularnewline
 & 706  & 669  & 669\tabularnewline
\hline 
keep [belts 0,3,4,5]  & 5,184 (129 KB)  & 1,440 (33 KB)  & 1,440 (33 KB)\tabularnewline
keep [bln\_analyzed 2,3] & 3.43  & 3.50  & 3.50\tabularnewline
 & 1,854  & 1,822  & 1,822\tabularnewline
\hline 
keep [belts 4,5]  & 2,304 (65 KB)  & 1,920 (65 KB)  & 1,920 (65 KB)\tabularnewline
keep [bln\_analyzed all] & 5.46  & 5.49  & 5.49\tabularnewline
 & 952  & 940  & 940\tabularnewline
\hline 
keep [belts 3,4,5]  & 1,728 (65 KB)  & 960 (33 KB)  & 960 (33 KB)\tabularnewline
keep [bln\_analyzed 3,4,5] & 2.75  & 2.87  & 2.87\tabularnewline
 & 2,151  & 2,100  & 2,100\tabularnewline
\hline 
keep [belts 0,1,2]  & 1,728 (65 KB)  & 960 (33 KB)  & 960 (33 KB)\tabularnewline
keep [bln\_analyzed 0,1,2]  & 2.75  & 2.87  & 2.87\tabularnewline
 & 2,149  & 2,098  & 2,098\tabularnewline
\hline 
keep [belts 1,3,5]  & 1,728 (65 KB)  & 960 (33 KB)  & 960 (33 KB)\tabularnewline
keep [bln\_analyzed 1,3,5]  & 2.73  & 2.79  & 2.79\tabularnewline
 & 2,159  & 2,134  & 2,134\tabularnewline
\hline

\caption{Projection abstractions of the 6-Belts Scanalyzer in the standard representation. This state space has 46,080 reachable original states with an average distance of 8.34.}\label{tab:SCNZ-ST-6}

\end{longtable} 

\end{footnotesize}


\begin{footnotesize}

\begin{longtable}{|l|l|l|l|}

\hline
\rowcolor{LightGrey} Abstraction  & Original  & Mutex  & True\tabularnewline
\hline
\rowcolor{LightGrey}  & Size  & Size  & Size\tabularnewline
\rowcolor{LightGrey}  & Avg.\ $h$  & Avg.\ $h$  & Avg.\ $h$\tabularnewline
\rowcolor{LightGrey}  & Avg.\ Nodes  & Avg.\ Nodes  & Avg.\ Nodes\tabularnewline
\hline
\endfirsthead
\multicolumn{4}{l}%
{\tablename\ \thetable\ -- \textit{Continued from previous page}} \\
\hline
\rowcolor{LightGrey} Abstraction  & Original  & Mutex  & True\tabularnewline
\hline
\rowcolor{LightGrey}  & Size  & Size  & Size\tabularnewline
\rowcolor{LightGrey}  & Avg.\ $h$  & Avg.\ $h$  & Avg.\ $h$\tabularnewline
\rowcolor{LightGrey}  & Avg.\ Nodes  & Avg.\ Nodes  & Avg.\ Nodes\tabularnewline
\hline
\endhead

$3\leftarrow blank$  & 239,500,800 (8.1 GB)  & 1,995,840 (65 MB)  & 1,995,840 (65 MB)\tabularnewline
 & 39.43  & 47.63  & 47.63\tabularnewline
 & 9,464  & 87  & 87\tabularnewline
\hline 
$11\leftarrow blank$  & 239,500,800 (8.1 GB)  & 2,540,160 (65 MB)  & 1,673,280 (65 MB)\tabularnewline
 & 31.94  & 31.94  & 32.75\tabularnewline
 & 73,298  & 73,298  & 62,073\tabularnewline
\hline 
$3\leftarrow blank,7$  & 79,833,600 (2.1 GB)  & 2,177,280 (65 MB)  & 1,491,840 (33 MB)\tabularnewline
 & 28.88  & 30.98  & 32.02\tabularnewline
 & 136,063  & 84,324  & 66,462\tabularnewline
\hline

\caption{Domain abstractions of the $3\times4$ Constrained-Movement Sliding-Tile Puzzle. This state space has 2,177,280 states with an average heuristic value of 47.71.}\label{tab:SP-SE-3by4}

\end{longtable} 

\end{footnotesize}


\begin{footnotesize}

\begin{longtable}{|l|l|l|l|}

\hline
\rowcolor{LightGrey} Abstraction  & Original  & Mutex  & True\tabularnewline
\hline
\rowcolor{LightGrey}  & Size  & Size  & Size\tabularnewline
\rowcolor{LightGrey}  & Avg.\ $h$  & Avg.\ $h$  & Avg.\ $h$\tabularnewline
\rowcolor{LightGrey}  & Avg.\ Nodes  & Avg.\ Nodes  & Avg.\ Nodes\tabularnewline
\hline
\endfirsthead
\multicolumn{4}{l}%
{\tablename\ \thetable\ -- \textit{Continued from previous page}} \\
\hline
\rowcolor{LightGrey} Abstraction  & Original  & Mutex  & True\tabularnewline
\hline
\rowcolor{LightGrey}  & Size  & Size  & Size\tabularnewline
\rowcolor{LightGrey}  & Avg.\ $h$  & Avg.\ $h$  & Avg.\ $h$\tabularnewline
\rowcolor{LightGrey}  & Avg.\ Nodes  & Avg.\ Nodes  & Avg.\ Nodes\tabularnewline
\hline
\endhead

$3\leftarrow13,blank$  & 54,432,000 (3.1 GB)  & 1,814,400 (97 MB)  & 1,310,400 (49 MB)\tabularnewline
 & 37.22  & 37.22  & 39.20\tabularnewline
 & 103,244  & 103,244  & 77,219\tabularnewline
\hline 
$17\leftarrow7,blank$  & 54,432,000 (3.1 GB)  & 1,814,400 (97 MB)  & 1,310,400 (49 MB)\tabularnewline
 & 36.90  & 36.90  & 38.92\tabularnewline
 & 106,600  & 106,600  & 78,511\tabularnewline
\hline 
\multicolumn{1}{c}{} & \multicolumn{1}{c}{} & \multicolumn{1}{c}{} & \multicolumn{1}{c}{} \tabularnewline
\multicolumn{1}{c}{} & \multicolumn{1}{c}{} & \multicolumn{1}{c}{} & \multicolumn{1}{c}{} \tabularnewline
\multicolumn{1}{c}{} & \multicolumn{1}{c}{} & \multicolumn{1}{c}{} & \multicolumn{1}{c}{} \tabularnewline
$7\leftarrow14,blank$  & 54,432,000 (3.1 GB)  & 1,814,400 (97 MB)  & 1,310,400 (49 MB)\tabularnewline
 & 36.82  & 36.82  & 39.07\tabularnewline
 & 107,252  & 107,252  & 75,748\tabularnewline
\hline

\caption{Domain abstractions of the $4\times5$ Constrained-Movement Sliding-Tile Puzzle. This state space has 1,814,400 states with an average heuristic value of 65.81.}\label{tab:SP-SE-4by5}

\end{longtable} 

\end{footnotesize}

\end{document}